%% file: main.tex
\documentclass[sigconf, screen, dvipsnames, authorversion]{acmart}

\usepackage[noend]{algorithmic}
\usepackage{algorithm}
\usepackage{multirow}
\usepackage{amsmath}
\usepackage{mathtools}
\usepackage[inline]{enumitem} % Inline lists
\usepackage{subcaption}
\usepackage{bm} % Bold greek
\usepackage{xcolor}

\usepackage{lipsum}

\usepackage{wrapfig}

\usepackage{arydshln}
\makeatletter
\def\adl@drawiv#1#2#3{
        \hskip.5\tabcolsep
        \xleaders#3{#2.5\@tempdimb #1{1}#2.5\@tempdimb}%
                #2\z@ plus1fil minus1fil\relax
        \hskip.5\tabcolsep}
\newcommand{\cdashlinelr}[1]{%
  \noalign{\vskip\aboverulesep
          \global\let\@dashdrawstore\adl@draw
          \global\let\adl@draw\adl@drawiv}
  \cdashline{#1}
  \noalign{\global\let\adl@draw\@dashdrawstore
          \vskip\belowrulesep}}
\makeatother

\usepackage{colortbl}
\newcolumntype{R}{S[table-format=1.2]}
\usepackage{siunitx}
\usepackage{tikz}
\usetikzlibrary{automata, arrows, bayesnet, bending}

\usepackage{tikzscale}

\copyrightyear{2024}
\acmYear{2024}
\setcopyright{acmlicensed}\acmConference[KDD '24]{Proceedings of the 30th ACM SIGKDD Conference on Knowledge Discovery and Data Mining}{August 25--29, 2024}{Barcelona, Spain}
\acmBooktitle{Proceedings of the 30th ACM SIGKDD Conference on Knowledge Discovery and Data Mining (KDD '24), August 25--29, 2024, Barcelona, Spain}
\acmDOI{10.1145/3637528.3671512}
\acmISBN{979-8-4007-0490-1/24/08}

\begin{document}
\title{Learning Metrics that Maximise Power for Accelerated A/B-Tests}

\author{Olivier Jeunen}
\affiliation{
  \institution{ShareChat}
  \city{Edinburgh}
  \country{United Kingdom}
} 
\email{jeunen@sharechat.co}

\author{Aleksei Ustimenko}
\affiliation{
  \institution{ShareChat}
  \city{London}
  \country{United Kingdom}
}
\email{aleksei.ustimenko@sharechat.co}

\begin{abstract}
Online controlled experiments are a crucial tool to allow for confident decision-making in technology companies.
A North Star metric is defined (such as long-term revenue or user retention), and system variants that statistically significantly improve on this metric in an A/B-test can be considered superior.
North Star metrics are typically delayed and insensitive.
As a result, the cost of experimentation is high: experiments need to run for a long time, and even then, type-II errors (i.e. \emph{false negatives}) are prevalent.

We propose to tackle this by \emph{learning} metrics from short-term signals that directly maximise the statistical power they harness with respect to the North Star.
We show that existing approaches are prone to overfitting, in that higher \emph{average} metric sensitivity does not imply improved type-II errors, and propose to instead minimise the $p$-values a metric would have produced on a log of past experiments.
We collect such datasets from two social media applications with over 160 million Monthly Active Users each, totalling over 153 A/B-pairs.
Empirical results show that we are able to increase statistical power by up to 78\% when using our learnt metrics stand-alone, and by up to 210\% when used in tandem with the North Star.
Alternatively, we can obtain constant statistical power at a sample size that is down to 12\% of what the North Star requires, significantly reducing the cost of experimentation.
\end{abstract}




\maketitle

\input{1_Intro}
\input{2_Background}
\input{3_Contributions}
\input{4_Experiments}
\input{5_Insights}
\input{6_Conclusions}

\bibliographystyle{ACM-Reference-Format}
\bibliography{bibliography}

\appendix
\input{A_Reproducibility}

\end{document}

%% file: 1_Intro.tex
% !TEX root = main.tex
\section{Introduction \& Motivation}
Modern platforms on the web need to continuously make decisions about their product and user experience, which are often central to the business at hand.
These decisions range from design and interface choices to back-end technology adoption and machine learning models that power personalisation.
Online controlled experiments, the modern web-based extension of Randomised Controlled Trials (RCTs)~\cite{Rubin1974}, provide an effective tool to allow for confident decision-making in this context~\cite{kohavi2020trustworthy} (bar some common pitfalls~\cite{Kohavi2022,Jeunen2023_Forum}).

A North Star metric is adopted, such as long-term revenue or user retention, and system variants that statistically significantly improve the North Star metric are considered superior to the tested alternative~\cite{Deng2016}.
Proper use of statistical hypothesis testing tools such as Welch's $t$-test~\cite{Welch1947}, then allows us to define and measure statistical significance in a mathematically rigorous manner.

However effective this procedure is, it is far from efficient.
Indeed, experiments typically need to run for a long time, and statistically significant changes to the North Star are scarce.
This can either be due to false negatives (i.e. type-II error), or simply because the North Star is not moved by short-term experiments.
In these cases, we need to resort to second-tier metrics (e.g. various types of user engagement signals) to make decisions instead.
These problems are common in industry, as evidenced by a wide breadth of related work.
A first line of research leverages control variates to reduce the variance of the North Star metric, directly reducing type-II errors by increasing sensitivity~\cite{Deng2013,Xie2016, Budylin2018, Poyarkov2016, Guo2021, Baweja2024}.
Another focuses on identifying second-tier ``\emph{proxy}'' or ``\emph{surrogate}'' metrics that are promising to consider instead of the North Star~\cite{Wang2022,Richardson2023,Tripuraneni2023}, or to predict long-term effects from short-term data~\cite{Athey2019,Tang2022,Goffrier2023}.
Finally, several works learn metric \emph{combinations} that maximise sensitivity~\cite{Deng2016,Kharitonov2017,Tripuraneni2023}.

This paper synthesises, generalises and extends several of the aforementioned works into a general framework to learn A/B-testing metrics that maximise the statistical power they harness.
We specifically extend the work of \citeauthor{Kharitonov2017}~\cite{Kharitonov2017} to applications beyond web search, where the North Star can be delayed and insensitive.
We highlight how their approach of maximising the average $z$-score does not accurately reflect downstream metric utility in our case, in that it does not penalise \emph{disagreement} with the North Star sufficiently (i.e. type-III/S errors~\cite{Mosteller1948,Kaiser1960,Gelman2014,Urbano2019}).
Indeed: whilst this approach maximises the \emph{mean} $z$-score, it does not necessarily improve the \emph{median} $z$-score, and does not lead to improved statistical power in the form of reduced type-II error as a result.

Alternatively, optimising the learnt metric to minimise $p$-values ---either directly or after applying a $\log$-transformation--- more equitably ditributes gains over multiple experiments, leading to \emph{more} statistically significant results instead of a few \emph{extremely} significant results.
Furthermore, we emphasise that learnt metrics are not meant to \emph{replace} existing metrics, but rather to complement them.
As such, their evaluation should be done through multiple hypothesis testing (with appropriate corrections~\cite{Shaffer1995}) if \emph{any} of the North Star, available vetted proxies and surrogates, \emph{or} learnt metrics are statistically significant under the considered treatment variant.
We can then either adopt a conservative plug-in Bonferroni correction to temper type-I errors, or analyse synthetic A/A experiments to ensure the final procedure matches the expected confidence level.

We empirically validate these insights through two dataset of past logged A/B results from large-scale short-video platforms with over 160 million monthly active users each: ShareChat and Moj.
Experimental results highlight that our learnt metrics provide significant value to the business: learnt metrics can increase statistical power by up to 78\% over the North Star, and up to 210\% when used in tandem.
Alternatively, if we wish to retain constant statistical power as we do under the North Star, we can do so with down to 12\% of the original required sample size.
This significantly reduces the cost of online experimentation to the business.
Our learnt metrics are currently used for confident, high-velocity decision-making across ShareChat and Moj business units.

%% file: 2_Background.tex
\section{Background \& Problem Setting}
We deal with online controlled experiments, where two system variants $A$ and $B$ are deployed to a properly randomised sub-population of users, adhering to best practices~\cite{kohavi2020trustworthy, Jeunen2023_Forum}.

For every system variant, for every experiment, we measure various metrics that describe how users interact with the platform.
These metrics include types of implicit engagement (e.g. video-plays and watch time), as well as explicit engagement (e.g. likes and shares) as well as longer-term retention or revenue signals.
For each metric, we log empirical means, variances and covariances (of the sample mean).
For metrics $m_{i}$ with $1\leq i \leq N$, that is:
\begin{equation}
\bm{\mu} = [\mu_{1}, \ldots \mu_{N}], \text{ and } \bm{\Sigma} = \begin{bmatrix}    \sigma_{1} & \dots & \sigma_{1i} & \dots & \sigma_{1N}\\
\vdots & \ddots & \vdots & \ddots & \vdots \\
\sigma_{i1} & \dots & \sigma_{i} & \dots & \sigma_{iN}\\
\vdots & \ddots & \vdots & \ddots & \vdots\\
\sigma_{N1} & \dots & \sigma_{Ni} & \dots & \sigma_{N}
\end{bmatrix}.    
\end{equation}
Superscripts denote measurements pertaining to different variants in an experiment: e.g. $\bm{\mu}^{A}$ and $\bm{\mu}^{B}$.

\subsection{Statistical Significance Testing}
We want to assess whether the mean of metric $m_{i}$ is statistically significantly higher under variant $A$ compared to variant $B$.
To this end, we define a significance level $\alpha$ (often $\alpha \approx 0.05$), corresponding to the false-positive-rate we deem acceptable.
Then, we apply Welch's $t$-test.
The test statistic (also known as the $z$-score) for metric $m_{i}$ and the given variants is given by: 
\begin{equation}
    z_{i}^{A\succ B} = \frac{\mu_{i}^{A} - \mu_{i}^{B}}{\sqrt{\sigma^{A}_{i} + \sigma^{B}_{i}}}.
\end{equation}
We then transform this to a $p$-value for a two-tailed test as:
\begin{equation}\label{eq:pvalue}
    p_{i}^{A\neq B} = 2\cdot \min (\Phi (z_{i}^{A\succ B});1-\Phi(z_{i}^{A\succ B})).
\end{equation}
Here, $A \succ B$ denotes a partial ordering between variants, implying that $A$ is preferred over $B$.
$\Phi(\cdot)$ represents the cumulative distribution function (CDF) for a standard Gaussian.
For completeness, this CDF is given by: 
\begin{equation}
    \Phi(z) = \frac{1}{\sqrt{2\pi}} \int_{-\infty}^{z} e^{-\frac{t^{2}}{2}}{\rm d}t.
\end{equation}
When $p_{i}^{A\neq B} < \alpha$, we can confidently \emph{reject} the null-hypothesis that $A$ and $B$ are equivalent w.r.t. the mean of metric $m_i$.
Note that $z$-scores are \emph{signed}, whereas two-tailed $p$-values are not.
Indeed: relabeling the variants changes the $z$-score but not the $p$-value, which leaves room for faulty conclusions of directionality, known as type-III errors~\cite{Mosteller1948,Kaiser1960,Urbano2019} or \emph{sign} errors~\cite{Gelman2014}.
We discuss these phenomena in detail, further in this article.

A one-tailed $p$-value for the one-tailed null hypothesis $A\nsucc B$ is given by $p_{i}^{A\nsucc B} = 1 - \Phi(z_{i}^{A \succ B})$, and rejected when $<\frac{\alpha}{2}$.
Throughout, we use two-tailed $p$-values unless mentioned otherwise.

\subsubsection{$p$-value corrections}
The above procedure is valid for a single metric, a single hypothesis, and importantly, a single decision.
Nevertheless, this is not how experiments run in practice.
Without explicit corrections on the $p$-values (or corresponding $z$-scores), violations of these assumptions lead to inflated false-positive-rates.
We consider two common cases: a (conservative) multiple testing correction when an experiment has several treatments, and a sequential testing correction when experiments have no predetermined end-date or sample size at which to conclude.
These corrections are applied as experiment-level corrections, to ensure that for any metric $m_{i}$ and variants $A,B$, the obtained $p$-values accurately reflect what they should reflect, yielding the specified coverage at varying confidence levels $\alpha$.

\paragraph{Multiple comparisons}\label{sec:pvalue_corrections}
Often, launched experiments will have multiple treatments deployed, leading to the infamous ``\emph{multiple hypothesis testing}'' problem~\cite{Shaffer1995}.
We apply a Bonferroni correction to deal with this.
When there are $T$ treatments, we consider a treatment to be statistically significantly different from control when a two-tailed $p <\frac{\alpha}{T}$, instead of the original $p < \alpha$ threshold.

We can equivalently apply this correction on $z$-scores instead, allowing us to directly compare $z$-scores across experiments with varying numbers of treatments.
Recall that the percentile point function is the inverse of the CDF.
We obtain a one-tailed $p$-value as $p = 1 - \Phi(z)$, and we reject the one-tailed null hypothesis when $p < \frac{\alpha}{2}$.
Now, instead, we reject when $p < \frac{\alpha}{2T}$.
As such, computing corrected $z$-scores as $\bar{z} = z \frac{\Phi^{-1}(\frac{\alpha}{2})}{\Phi^{-1}(\frac{\alpha}{2T})}$ controls type-I errors effectively.

\paragraph{Always-Valid-Inference and peeking}
A statistical test should only be performed \emph{once}, at the end of an experiment.
When the treatment effect is large, this implies we may have been able to conclude the experiment earlier.
To this end, \emph{sequential} hypothesis tests have been proposed in the literature~\cite{Wald1945}.
Modern versions make use of Always-Valid-Inference (AVI)~\cite{Howard2021} to allow for continuous \emph{peeking} at intermediate results and making decisions based on them, whilst controlling type-I errors.
Here, analogously, we can apply a correction on the $z$-scores as follows:
\begin{equation}
\bar{z} = \frac{z}{\sqrt{\frac{(N_{AB}+\rho)}{N_{AB}} \log\left( \frac{N_{AB}+\rho}{\rho\alpha^{2}}\right)}}, \quad \rho = \frac{10\,000}{\log(\log(\frac{e}{\alpha^{2}}))-2\log(\alpha)},
\end{equation}
where $N_{AB}$ is the total number of samples over both variants combined.
For a detailed motivation, see \citeauthor{schmit2022sequential}~\cite{schmit2022sequential}.

These corrections are applied on a per-experiment level, both in the objective functions of methods introduced in the following Sections and when evaluating the metrics that they produce.

\subsection{Learning Metrics that Maximise Sensitivity}
The observation that we can learn parameters to maximise statistical sensitivity is not new.
\citeauthor{Yue2010} apply such ideas specifically for interleaving experiments in web search~\cite{Yue2010}.
\citeauthor{Kharitonov2017} extend this to A/B-testing in web search, aiming to learn combinations of metrics that maximise the average $z$-score~\cite{Kharitonov2017}.
\citeauthor{Deng2016} discuss ``lessons learned'' from applying similar techniques~\cite{Deng2016}.
We introduce the approach presented by \citeauthor{Kharitonov2017}~\cite{Kharitonov2017}, as our proposed improvements build on their foundations.

We consider new metrics as linear transformations on $\bm{\mu}$: 
\begin{equation}
    \omega = \bm{\mu} w^{\intercal}, \text{ where } w \in \mathbb{R}^{1\times N}.    
\end{equation}
The advantage of restricting ourselves to linearity, it that we can write out the $z$-score of the new metric as a function of its weights:
\begin{equation}\label{eq:zscore_learnt}
z^{A\succ B}_{\omega} = \frac{\bm{\mu}^{A}w^{\intercal}-\bm{\mu}^{B}w^{\intercal}}{\sqrt{w\bm{\Sigma}^{A}w^{\intercal}+w\bm{\Sigma}^{B}w^{\intercal}}}.    
\end{equation}
These $z$-scores can be used exactly as before to obtain $p$-values.
An intuitive property of the $z$-score, is that a relative $z$-score of $r = \frac{z^{A\succ B}_{\omega}}{z^{A\succ B}_{i}}$ implies that $\omega$ requires a factor $r^{2}$ fewer samples to achieve the same significance level as $m_{i}$~\cite{Chapelle2012}.
This can directly be translated to the cost of experimentation, as it allows us to run experiments for shorter time-periods or on smaller sub-populations.

As such, it comes naturally to frame the objective as learning the weights $w$ that maximise the $z$-score on the training data.
This training data consists of a set of experiments with pairs of variants $\mathcal{E} = \{(A,B)_{i}\}_{i=1}^{|\mathcal{E}|}$.
We consider three distinct relations between pairs of deployed system variants:
\begin{enumerate}
    \item \emph{Known} outcomes: $(A,B) \in\mathcal{E}^{+}$, where $A \succ B$, 
    \item \emph{Unknown} outcomes: $(A,B) \in\mathcal{E}^{?}$, where $A~?~B$,  %  \stackrel{?}{\succ}
    \item \emph{A/A} outcomes: $(A,B) \in\mathcal{E}^{\simeq}$, where $A \simeq B$.
\end{enumerate}
Here, $A\succ B$ implies that there is a \emph{known} and vetted preference of variant $A$ over $B$ --- typically because the North star or other guardrail metrics showed statistically significant improvements.
These experiments are further validated by replicating outcomes, observing long-term holdouts, or because the experiment was part of an intentional degradation test. 
We denote \emph{inconclusive} experiments as $A~?~B$, implying statistically insignificant outcomes on the North Star.
In rare cases, the North star might have gone up at the expense of important guardrail metrics, rendering conclusions ambiguous.
We only include experiments into the inconclusive set for which we have a very strong intuition that something changed (and we ``know'' the null hypothesis should be rejected), but we are unable to make a confident directionality decision. 
This ensures that we can use this set to truly measure type-II errors.
Finally, $A \simeq B$ represents A/A experiments, where we know the null hypothesis to hold true (by design).
The first set of experiments is used to measure type-III/S errors.
Known and unknown outcomes are used to measure type-II errors, and A/A experiments can inform us about type-I errors.
This dataset of past A/B experiments is collected and labelled by hand, from natural experiments occurring on the platform over time.

\subsubsection{Optimising Metric Weights with a Geometric Heuristic}\label{sec:heuristic}
Note that $z$ as a function of $w$ is \emph{scale-free}. That is, the \emph{direction} of the weight vector $w$ matters, but its scale does not. 
As \citeauthor{Kharitonov2017} write~\cite{Kharitonov2017}, we can compute the optimal direction of $w$ using the Lagrange multipliers method, to obtain:
\begin{equation}
w^{\star}_{A\succ B}\propto (\bm{\Sigma}^{A}+\bm{\Sigma}^{B}+\epsilon I)^{-1}(\bm{\mu}^{A}-\bm{\mu}^{B}).
\end{equation}
Here, $\epsilon \in \mathbb{R}$ is a small number to ensure that the matrix to be inverted is not singular.
\citeauthor{Kharitonov2017} fix this value at $\epsilon = 0.01$ and never adjust it throughout the paper.
We wish to highlight that technique is known as Ledoit-Wolf shrinkage~\cite{Ledoit2004,Ledoit2020}, and that it can have \emph{substantial} influence on the obtained direction.
Indeed: it acts as a regularisation term pushing the weights closer to $w =(\bm{\mu}^{A}-\bm{\mu}^{B})$.
This can be seen by observing that as $\epsilon \to \inf$, the inverse becomes $\frac{1}{\epsilon}I$, and the solution hence becomes $w = \frac{(\bm{\mu}^{A}-\bm{\mu}^{B})}{\epsilon}$.
As we only care about the direction we can ignore the denominator.
To ensure fair comparison, we also set $\epsilon = 0.01$.
Exploring the effects of Ledoit-Wolf shrinkage as a regularisation technique where $\epsilon$ is a hyper-parameter, gives an interesting avenue for future work.

In order to include observations from multiple experiments into a single set of learned weights, they propose to compute the optimal direction per experiment, normalise, and average the weights:
\begin{equation}
w^{\star}_{\mathcal{E}^{+}} = \frac{1}{|\mathcal{E}^{+}|}   \sum_{(A,B) \in\mathcal{E}^{+}} \frac{w^{\star}_{A \succ B}}{\left\Vert w^{\star}_{A \succ B}\right\Vert_{2}}.    
\end{equation}
Whilst this procedure provides no guarantees about the sensitivity of the obtained metric on the overall set of experiments, it is efficient to compute and provides a strong baseline method. 

\subsubsection{Optimising Metric Weights via Gradient Ascent on $z$-scores}
A more principled approach is to cast the above as an optimisation problem.
The objective function for this optimisation problem consists of three parts.
First, we wish to \emph{maximise} the $z$-score for all variant pairs with \emph{known} outcomes:
\begin{equation}    
\begin{split}
\mathcal{L}_{z}^{+}(w;\mathcal{E}^{+}) = \frac{1}{|\mathcal{E}^{+}|}   \sum_{(A,B) \in\mathcal{E}^{+}}  z^{A\succ B}_{\omega}.
\end{split}
\end{equation}
Second, we wish to \emph{maximise} the \emph{absolute} $z$-score for all variant pairs with inconclusive outcomes under the North Star:
\begin{equation}
\mathcal{L}_{z}^{?}(w;\mathcal{E}^{?}) = \frac{1}{|\mathcal{E}^{?}|}   \sum_{(A,B) \in\mathcal{E}^{?}} \left| z^{A\succ B}_{\omega}\right|. 
\end{equation}
Third, we wish to \emph{minimise} the \emph{absolute} $z$-score for all variant pairs that are equivalent (i.e. A/A-pairs):
\begin{equation}\label{eq:minimise_AA}
\mathcal{L}_{z}^{\simeq}(w;\mathcal{E}^{\simeq}) = \frac{1}{|\mathcal{E}^{\simeq}|}   \sum_{(A,B) \in\mathcal{E}^{\simeq}} \left| z^{A\succ B}_{\omega}\right|.
\end{equation}
This gives rise to a combined objective as a weighted average:
\begin{equation}
\mathcal{L}_{z}(w;\mathcal{E}) = \mathcal{L}_{z}^{+}(w;\mathcal{E}^{+})  + \lambda^{?}\mathcal{L}_{z}^{?}(w;\mathcal{E}^{?})  - \lambda^{\simeq}\mathcal{L}_{z}^{\simeq}(w;\mathcal{E}^{\simeq}) .    
\end{equation}
\citeauthor{Kharitonov2017} demonstrate that, for a variety of different metrics in a web search engine, these approaches can exhibit improved sensitivity~\cite{Kharitonov2017}.
We apply this method to learn instantaneously available proxies to a delayed North Star metric in general scenarios, and propose several extensions, detailed in the following Sections.

%% file: 3_Contributions.tex
\section{Methodology \& Contributions}
\subsection{Learning Metrics that Maximise Power}

When directly optimising $z$-scores, an implicit assumption is made that the utility we obtain from increased $z$-scores is linear.
This is seldom a truthful characterisation of reality, considering how we wish to actually \emph{use} these metrics downstream.
We provide a toy example in Table~\ref{tab:example}, reporting $z$-scores and one-tailed $p$-values for two experiments and three possible metrics $m_1, m_2, m_3$, inspired by real data.
In this toy example, we know that $A \succ B$, based on a hypothetical North Star metric. 
Nevertheless, as we do not know this beforehand, we typically test for the null hypothesis $A \simeq B$ with two-tailed $p$-values.
In the table, this means that the outcome is statistically significant if the reported one-tailed values are $p < \frac{\alpha}{2}$ \emph{or} $p > 1 - \frac{\alpha}{2}$.
We report the \emph{power} of every metric at varying significance levels $\alpha \in \{0.05, 0.01\}$, reporting whether
\begin{enumerate*}[label=(\roman*)]
    \item the null hypothesis is correctly rejected ($p < \frac{\alpha}{2}$, \textcolor{PineGreen}{\checkmark} ),
    \item the outcome is inconclusive (\textcolor{Orange}{\textbf{?}}), i.e. a type-II error, or
    \item the null hypothesis is rejected, but for the \emph{wrong reason} ($p > 1 - \frac{\alpha}{2}$, \textcolor{Maroon}{$\times$}).
\end{enumerate*}

This latter case is deeply problematic, as it signifies \emph{disagreement} with the North Star.
Such errors have been described as \emph{type-III} or \emph{type-S} errors in the statistical literature~\cite{Mosteller1948,Kaiser1960,Gelman2014,Urbano2019}.
Naturally, we would rather have a metric that fails to reject the null than one that confidently declares a faulty variant to be superior.
Indeed, \citeauthor{Deng2016} argue that both \emph{directionality} and \emph{sensitivity} are desirable attributes for any metric~\cite{Deng2016}.
Nevertheless, considering candidate metrics in Table~\ref{tab:example}, type-III errors are not sufficiently penalised by the average $z$-score: metric $m_{3}$ maximises this objective despite yielding statistical power that is on par with a coin flip.

\emph{Directly} maximising power might prove cumbersome, as it is essentially a discrete step-function w.r.t. the $z$-score, dependent on the significance level $\alpha$.
Instead, it comes natural to minimise the one-tailed $p$-value reported in Table~\ref{tab:example}.
Indeed, the $p$-value transformation models \emph{diminishing returns} for high $z$-scores, which allows type-III errors to be sufficiently penalised.
When considering this objective, $m_{3}$ is clearly suboptimal whilst $m_{2}$ is preferred.

Note that the change in objective would not affect the geometric heuristic as described in Section~\ref{sec:heuristic}.
As we simply apply a monotonic transformation on the $z$-scores, the weight direction that maximises the $z$-score equivalently minimises its $p$-value.
When learning via gradient descent, however, the $p$-value transformation affects how we aggregate and attribute gains over different input samples.
This allows us to stop focusing on increasing sensitivity for experiments that are already ``sensitive enough'', and more equitably consider all experiments in the training data.

\begin{table}[t]%\vspace{-3ex}
    \centering
    \begin{tabular}{lrcRlcc}
    \toprule
    ~&~&~& \textbf{$z$-score} & \textbf{$p$-value} & \multicolumn{2}{c}{\textbf{Power}} \\
    ~&~&~&~&~& $\alpha=0.05$ & $\alpha=0.01$ \\
    \midrule
    \multirow{2}{*}{$m_{1}$} &  \textbf{Exp. 1} &~& 1.97 & \num{2.44e-02} & \textcolor{PineGreen}{\checkmark}  & \textcolor{Orange}{\textbf{?}} \\
    ~&  \textbf{Exp. 2} &~& 1.97 & \num{2.44e-02} & \textcolor{PineGreen}{\checkmark}  & \textcolor{Orange}{\textbf{?}} \\
    ~&  \textit{Mean} &~& 1.97 & \num{2.44e-02} & \textbf{100\%} & 0\% \vspace{0.35ex}\\
    \cdashline{1-7}
    ~\vspace{-2ex}\\
    \multirow{2}{*}{$m_{2}$} &  \textbf{Exp. 1} &~& 1.90 & \num{2.87e-02} & \textcolor{Orange}{\textbf{?}}  & \textcolor{Orange}{\textbf{?}} \\
    ~&  \textbf{Exp. 2} &~& 3.50 & \num{2.33e-04} & \textcolor{PineGreen}{\checkmark} & \textcolor{PineGreen}{\checkmark}\\
    ~&  \textit{Mean} &~& 2.70 & {\boldmath{\num{1.45e-02}}} & 50\% & \textbf{50\%} \vspace{0.35ex}\\
    \cdashline{1-7}
    ~\vspace{-2ex}\\
    \multirow{2}{*}{$m_{3}$} &  \textbf{Exp. 1} &~& -2.58 & \num{9.95e-01} & \textcolor{Maroon}{$\times$}  &\textcolor{Maroon}{$\times$}  \\
    ~&  \textbf{Exp. 2} &~& 8.00 & \num{6.66e-16} & \textcolor{PineGreen}{\checkmark}  & \textcolor{PineGreen}{\checkmark} \\
    ~&  \textit{Mean} &~& \textbf{5.29} & \num{4.98e-01} & 50\% & 50\%\\        
    \bottomrule
    \end{tabular}
    \caption{An example setting highlighting that maximising average $z$-scores ($m_{1}$) does not imply improved statistical power.
    Statistical power is dependent on the significance level $\alpha$, $p$-values provide a good proxy to optimise for.
    We observe similar trade-offs from real past A/B experiments.}
    \label{tab:example}
\end{table}

This change in objective provides an intuitive and efficient extension to existing approaches, allowing us to directly optimise the confidence we have to \emph{correctly} reject the null-hypothesis.
For known outcomes, the loss function is given by:
\begin{equation}
    \begin{split}
\mathcal{L}_{p}^{+}(w;\mathcal{E}^{+}) = \frac{1}{|\mathcal{E}^{+}|}   \sum_{(A,B) \in\mathcal{E}^{+}} 1-\Phi( z^{A\succ B}_{\omega}) \\
= \frac{1}{|\mathcal{E}^{+}|}   \sum_{(A,B) \in\mathcal{E}^{+}} 1 - \Phi\left(\frac{\bm{\mu}^{A}w^{\intercal}-\bm{\mu}^{B}w^{\intercal}}{\sqrt{w\bm{\Sigma}^{A}w^{\intercal}+w\bm{\Sigma}^{B}w^{\intercal}}}\right),
    \end{split}
\end{equation}
and analogously extended to unknown $\mathcal{L}_{p}^{?}$ and A/A-outcomes $\mathcal{L}_{p}^{\simeq}$.
Nevertheless, we wish to point out that we only want to maximise $p$-values for A/A-outcomes if type-I error becomes problematic.
As we will show empirically in Section~\ref{sec:power_exp}, this is not a problem we encounter.
For this reason, we set $\lambda^{\simeq} \equiv 0$.

\begin{figure}%\vspace{-3ex}
    \centering
    \includegraphics[width=0.8\linewidth]{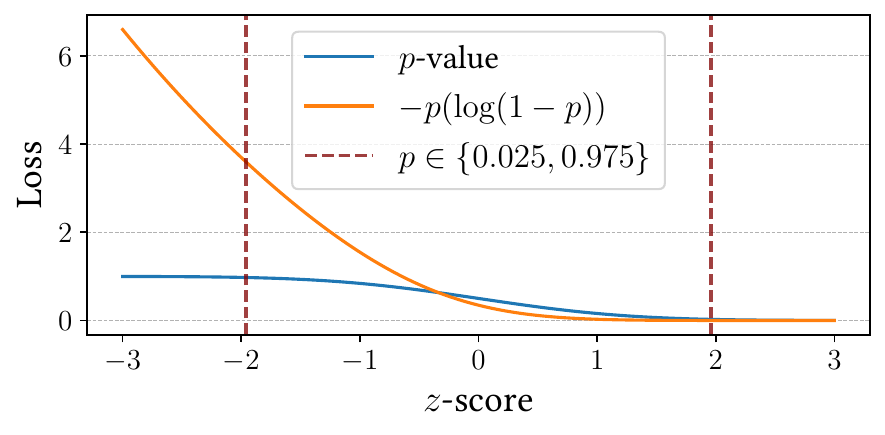}
    \caption{Visualising our proposed optimisation objectives for learnt metrics, as a function of their $z$-score.}
    \label{fig:losses}
\end{figure}

Note that whilst direct optimisation of $p$-values is an improvement over myopic consideration of $z$-scores, there is another caveat: the “worst-case” loss of a type-III/S error is bounded at 1, which does not reflect our true utility function: metrics that disagree with the North Star are \emph{far} less reliable than those that simply remain inconclusive.
As such, we also consider another variant of the objective, where  $\bar{p} =-p\log(1-p)$.
Figure~\ref{fig:losses} provides visual intuition to clarify how this monotonic transformation on the $p$-values more heavily penalises type-III/S errors, whilst retaining the optimum.
From a theoretical perspective, this function provides a convex relaxation for minimising the number of type-III/S errors a metric produces.
As a result, we expect this surrogate to exhibit strong generalisation.
We refer to this objective as minimising the $\log p$-value.

Note that one could envision further extensions here where the significance level $\alpha$ is \emph{directly} incorporated into the objective function to maximise statistical power at a given significance level $\alpha$.
Nevertheless, we conjecture that their discrete nature might hamper effective optimisation and generalisation, compared to the strictly convex and smooth surrogate we obtain from the $\log p$-value.

\subsection{Accelerated Convergence for Scale-Free Objectives via Spherical Regularisation}\label{sec:accelerate}
The objective functions we describe ---either $z$-scores or $(\log)p$-values--- are scale-free w.r.t. the weights that are being optimised.
As a result, out-of-the-box gradient-based optimisation techniques are not well-equipped to handle this efficiently.

Consider a simple toy example where we have two observed metrics for an experiment with a known preference $A \succ B$, and:
\begin{equation}
\bm{\mu}^{A} = [1.0, 1.0], \qquad \bm{\mu}^{B} = [0.5, 0.5],  \qquad \bm{\Sigma}^{A} = \bm{\Sigma}^{B} = I.
\end{equation}

\begin{figure*}[t!]
%\vspace{-3ex}
    \centering
    \begin{subfigure}[t]{0.5\textwidth}
        \centering
        \includegraphics[width=\linewidth]{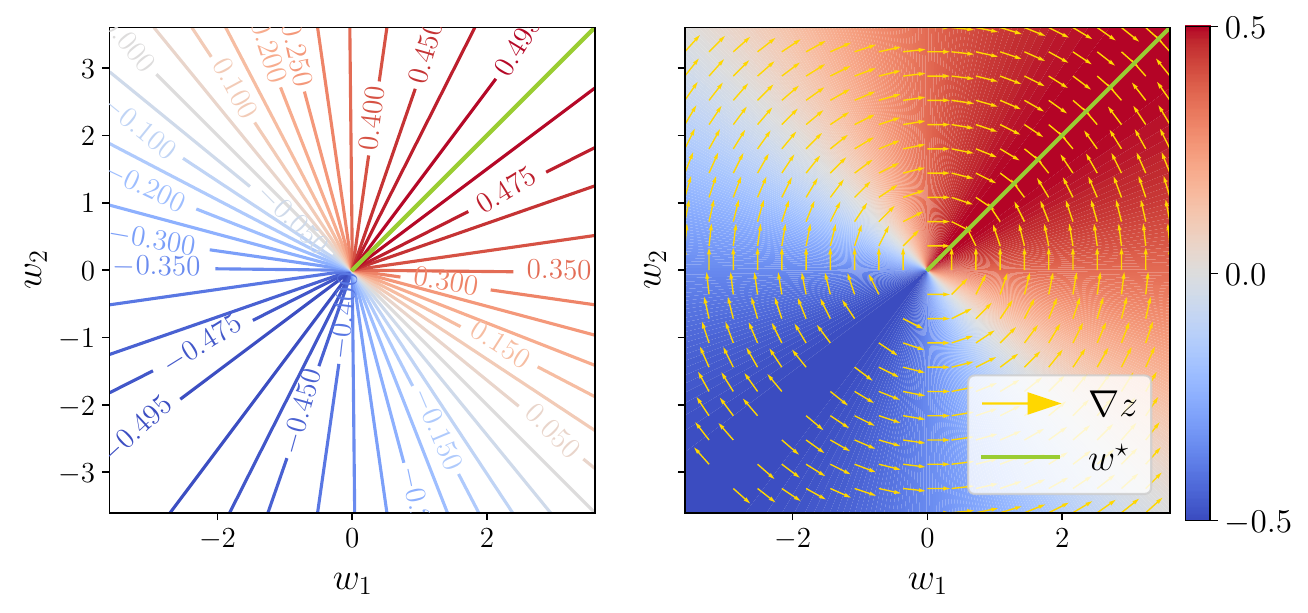}
        \caption{Original scale-free objective function.}\label{fig:orig_loss}
    \end{subfigure}
    ~ 
    \begin{subfigure}[t]{0.5\textwidth}
        \centering
        \includegraphics[width=\linewidth]{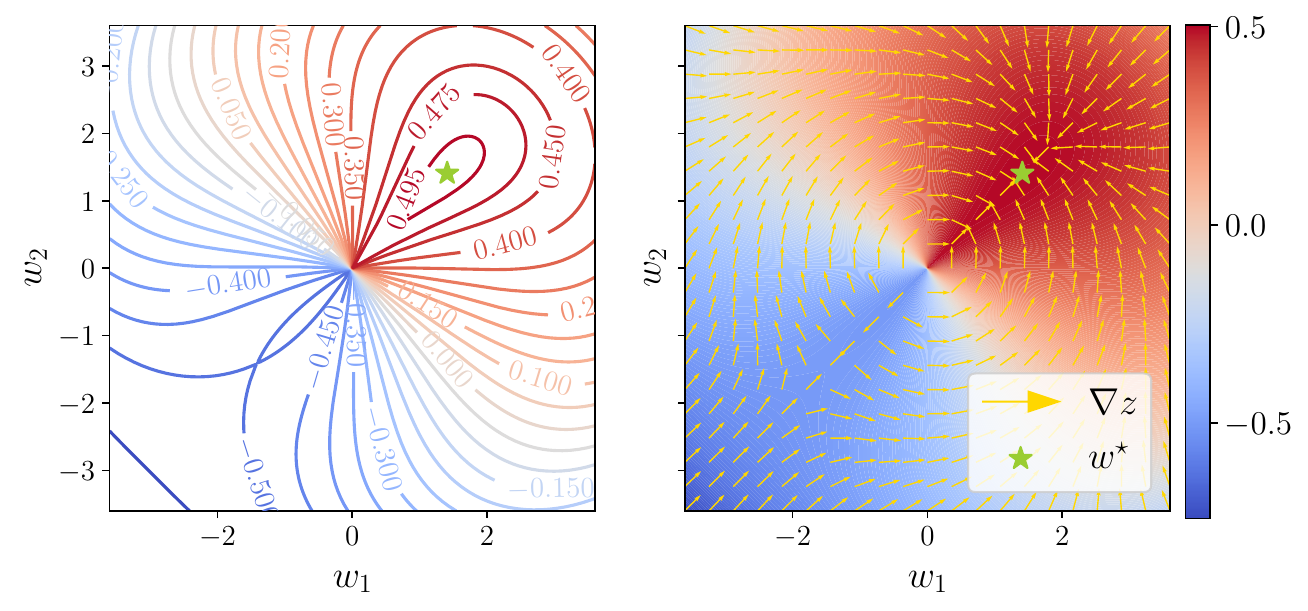}
        \caption{With added spherical regularisation ($\delta=\num{5e-4}$ ).}\label{fig:regularised_loss}
    \end{subfigure}
    \caption{
    Directly maximising $z$-scores yields a scale-free objective that is not amenable to efficient optimisation with gradient ascent (\emph{left}).
    Spherical regularisation retains all optima, whilst providing a gradient direction that is more aligned (\emph{right}).
    }\label{fig:regularisation_effect}
\end{figure*}

For this low-dimensional problem, we can visualise the $z$-score as a function of the metric weights in a contour plot, as shown in Figure~\ref{fig:orig_loss}.
Here, it becomes visually clear that whilst the \emph{direction} of the $w=[w_{1},w_{2}]$ vector matters, its scale does not.
The consequence is that the gradient vectors w.r.t. the objective on the right-hand plot can lead to slow convergence, even in this concave objective.
Indeed, for poor initialisation in the bottom left quadrant (e.g. $w = [-1, -2]$), the gradient direction is perpendicular w.r.t. the optima.

Recent work makes a similar observation for discrete scale-free objectives as they appear in ranking problems~\cite{Ustimenko2020}.
They propose to adopt projected gradient descent, normalising the gradients before every update.
Whilst effective, in our setting we would prefer to use out-of-the-box optimisation methods for practitioners' ease-of-use.
Instead, we introduce a simple regularisation term that represents the distance between the scale of the $w$ vector and a hyper-sphere:
\begin{equation}\mathcal{L}_{\lVert w \rVert} = -\delta \left(N - \lVert w\rVert_{2}^{2} \right)^{2}.\end{equation}
All optima for this objective function are also optima to the original function---but the gradient field is more amenable to iterative gradient-based optimisation techniques.
Figure~\ref{fig:regularised_loss} visualises how this transforms the loss surface.
Under this regularised objective, it is visually clear that gradient-based optimisation methods are likely to exhibit faster convergence. 
Our empirical results confirm this, for a variety of initialisation weights and learning objectives.

%% file: 4_Experiments.tex
\section{Experiments \& Discussion}\label{sec:experiments}
To empirically validate the methods proposed in this work, we require a dataset containing logged metrics (sample means, their variances and covariances), together with preference orderings over competing system variants that were collected from real-world A/B-tests, ideally spanning large user-bases and several weeks.

Existing work on this topic used a private dataset from Yandex focused on web search experiments that ran between 2013--2014~\cite{Kharitonov2017}.
They report type-I and -II errors for 8 metrics and a fixed 5\% significance level, over 118 A/B-tests and 472 A/A-tests.

In this work, we consider more general metrics that are relevant for use-cases beyond web search (i.a. user retention and various engagement signals).
Furthermore, we report type-I/II/III/S errors at varying significance levels, providing insights into the learnt metrics' behaviour.
For this, we leverage logs of past A/B-experiments on two large-scale short-video platforms with over 160 million monthly active users each: ShareChat and Moj. 
The datasets consist of 153 A/B-experiments (of which 58 were conclusive) total that ran in 2023, and over 25\,000 A/A-pairs.
In total, we have access to roughly 100 metrics detailing various types of interactions with the platform, engagements, and delayed retention signals.
Because our dataset is limited in size (a natural consequence of the problem domain), we are bound to overfit when using all available metrics as input features.
As such, we limit ourselves to 10 input metrics to learn from, and evaluate them w.r.t. the delayed North Star.
This feature selection step also ensures that our linear model consists of fewer parameters, which increases practitioners' and business stakeholders' trust in its output.
We focus on non-delayed signals, including activity metrics such as the number of sessions and active days, and counters for positive and negative feedback engagements of various types.
These are selected through an analysis of their type-I/II/III/S errors w.r.t. the North Star, as well as their $z$-scores: focusing on metrics with high sensitivity and limited disagreement.
The research questions we wish to answer empirically using this data, are the following:
\begin{description}
    \item[RQ1] \textit{Do learnt metrics effectively improve on their objectives?}
    \item[RQ2] \textit{How do learnt metrics behave in terms of  type-III/S errors?}
    \item[RQ3] \textit{How do learnt metrics' type-I/II errors behave when considered as stand-alone evaluation metrics?}
    \item[RQ4] \textit{How do learnt metrics' type-I/II errors behave when used in conjunction with the North Star and top proxy metrics?}
    \item[RQ5] \textit{How do learnt metrics influence required sample sizes when used in conjunction with the North Star and top proxy metrics?}
    \item[RQ6] \textit{Do we observe accelerated convergence over varying objectives via the proposed spherical regularisation technique?}
\end{description}

We report results for the ShareChat platform in what follows, and provide further empirical results for Moj in Appendix~\ref{sec:appendix}.

\subsection{Effectiveness of Learnt Metrics (RQ1)}
We learn and evaluate metrics through leave-one-out cross-validation: for every experiment, we train a model on all other experiments and evaluate the $z$-score (Eq.~\ref{eq:zscore_learnt}) and $p$-value (Eq.~\ref{eq:pvalue}) the metric yields for the held-out experiment.
We report the mean and median $z$-scores and $p$-values we obtain for all A/B-pairs with \emph{known} outcomes (i.e. $\mathcal{E}^{+}$) in Table~\ref{tab:objectives}.
Best performers for every column (either maximising $z$-scores or minimising $p$-values) are highlighted in \textbf{boldface}.
Empirical observations match our theoretical expectations: whilst the $z$-score objective does effectively maximise the \emph{average} $z$-score, it is the worst performer for both mean and median $p$-values, and even the median $z$-score.
Our proposed $\log p$-value objective effectively improves both the median $z$-score and $p$-value over alternatives.
\begin{table}[t]
    \centering
    \begin{tabular}{lcRRcll}
    \toprule
    \textbf{Objective} &~& \multicolumn{2}{R}{\textbf{$z$-score $\text{(}\uparrow{)}$}} &~& \multicolumn{2}{R}{\textbf{$p$-value $\text{(}\downarrow{)}$}} \\
    ~ &~ & $\text{Mean}$ & $\text{Median}$ &~& $\text{Mean}$ & $\text{Median}$ \\
    \midrule
    heuristic &~& 7.31 & 3.07 &~& \num{1.88e-01} & \num{1.18e-03} \\
    $z$-score  &~& {\bfseries 7.55} & 2.67 &~& \num{2.33e-01} & \num{3.88e-03} \\
    $p$-value &~& 5.22 & 3.08 &~& {\boldmath\num{4.32e-02}} & \num{1.09e-03} \\
    $\log p$-value &~& 4.33 & {\bfseries 3.17} &~& \num{5.19e-02} & {\boldmath\num{8.60e-04}} \\
    \bottomrule
    \end{tabular}
    \caption{Sensitivity results for learnt metrics, computed via leave-one-out cross-validation on all experiments with \emph{known} outcomes. We observe that minimising the $\log p$-value effectively improves median sensitivity over alternatives.}
    \label{tab:objectives}
\end{table}

\begin{figure*}[t!]
%\vspace{-3ex}
    \centering
    \includegraphics[width=\linewidth]{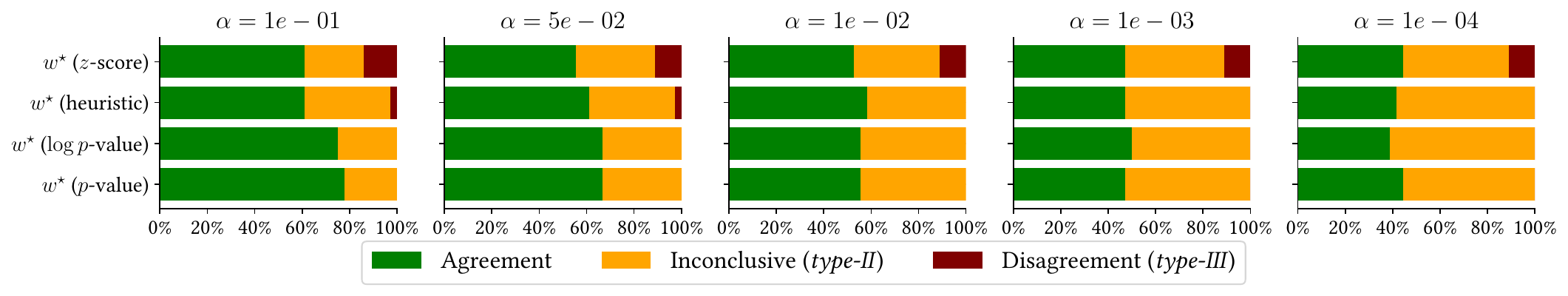}
    \caption{Learnt metrics and their agreement with \emph{known} North Star outcomes over varying significance levels. A learnt metric that maximises the average $z$-score exhibits significant type-III/S error, which can be alleviated by minimising $p$-values instead.}
    \label{fig:agreement}
\end{figure*}
\begin{figure*}[t!]
%\vspace{-3ex}
    \centering
    \begin{subfigure}[t]{\textwidth}
        \centering
        \includegraphics[width=\linewidth]{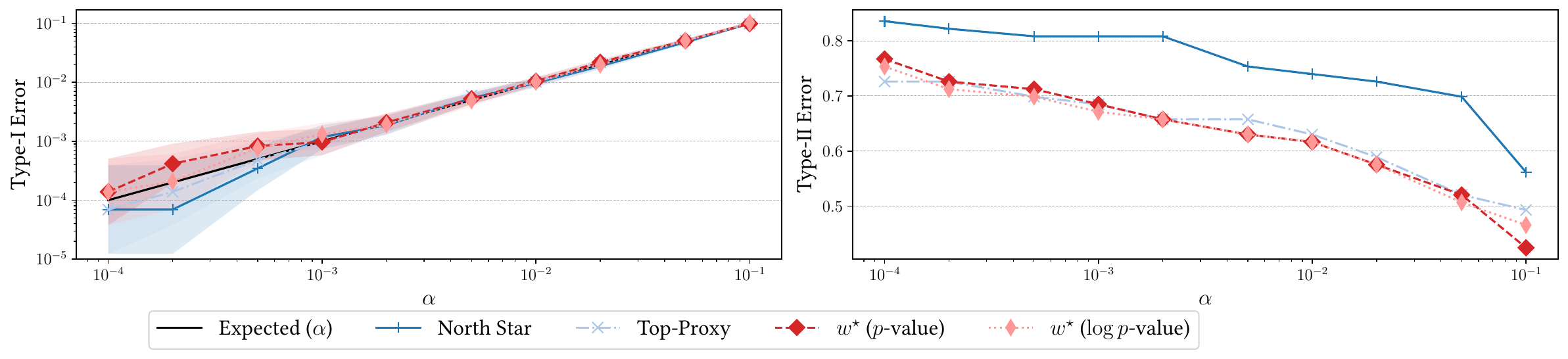}
        \caption{False-positive rate (type-I error) and false-negative rate (type-II error, $1-$ power) for stand-alone metrics.}
        \label{fig:power_standalone}
    \end{subfigure}
    ~\\
    \begin{subfigure}[t]{\textwidth}
        \centering
        \includegraphics[width=\linewidth]{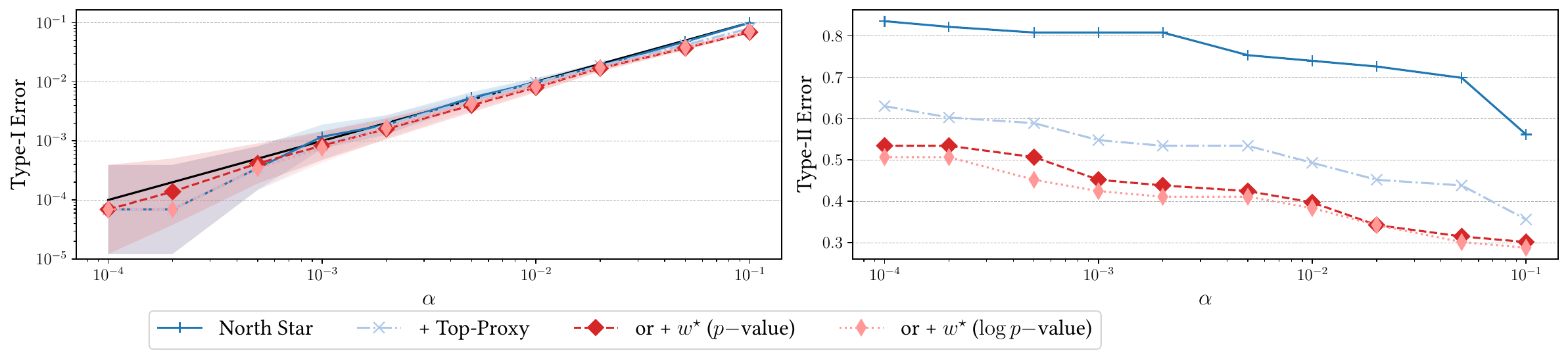}
        \caption{False-positive rate (type-I error) and false-negative rate (type-II error, $1-$ power) for sets of metrics, after Bonferroni correction.}
        \label{fig:power_combined}
    \end{subfigure}
    \caption{
    When considering \emph{only} learnt metrics, we improve type-II error significantly without hurting specificity.
    At $\alpha=0.05$, we increase statistical power by 67\%  (\emph{upper plot}).
    In \emph{conjunction} with the North Star and top proxy metrics, Bonferroni corrections are slightly conservative (type-I error $< \alpha$), and allow us to improve statistical power by 133\% for $\alpha=0.05$ (\emph{lower plot}).
    }\label{fig:power}
\end{figure*}
\subsection{Agreement with the North Star (RQ2)}
From the obtained $z$-scores and $p$-values summarised in Table~\ref{tab:objectives}, we can additionally derive (dis-)agreement with the North Star, for varying significance levels $\alpha$.
We visualise these results in Figure~\ref{fig:agreement}: if the obtained $p$-value under a learnt metric is lower than $\alpha$, that metric yields a statistically significant result (agreement).
If the obtained $p$-value for the alternative hypothesis (i.e. $B \succ A$ when we know $A \succ B$) is lower than $\alpha$, we have statistically significant \emph{disagreement}, or type-III error.
This is a capital sin we wish to avoid at all costs, as it severely diminishes the trust we can put in the learnt metric.
If the $p$-value reveals a statistically insignificant result, we say the result is inconclusive, implying a type-II error.
We observe that both the $z$-score-maximising metric and the heuristic approach fail to steer clear from type-III error. 
Optimising ($\log$)$p$-values instead, alleviates this issue.
For this reason, we only consider these metrics for further evaluation.
Indeed: an analysis of type-II error is rendered meaningless when type-III errors are present.

Figure~\ref{fig:agreement_Moj} in Appendix~\ref{sec:appendix} highlights that for the Moj platform as well, type-III errors are common in the case of $z$-score-maximising or heuristic metrics.
As such, we only consider ($\log$)$p$-values to assess increases in statistical power and potential reductions to the cost of running online experiments.

\subsection{Power Increase from Learnt Metrics (RQ3--4)}\label{sec:power_exp}
Until now, we have leveraged experiments with \emph{known} outcomes to assess sensitivity and agreement with the North Star.
Now, we additionally consider A/A-experiments ($\mathcal{E}^{\simeq}$) and experiments with \emph{unknown} outcomes ($\mathcal{E}^{?}$) to measure type-I and type-II error respectively.
We measure this for the North Star, for the best-performing proxy metric that serves as input to the learnt metrics, and for learnt metrics that exhibit no empirical disagreement with the North Star.
We plot the type-I error (i.e. fraction of A/A-pairs in $\mathcal{E}^{\simeq}$ that are statistically significant at significance level $\alpha$) and the type-II error (i.e. fraction of A/B-pairs in $\{\mathcal{E}^{+}\cup\mathcal{E}^{?}\}$ that are statistically insignificant at significance level $\alpha$) for varying values of $\alpha$ in Figure~\ref{fig:power_standalone}.
We observe that we are able to significantly reduce type-II errors compared to the North Star (up to 78\%), whilst keeping type-I errors at the required level (i.e. $\alpha$).
However, we also observe that the type-II error we obtain when using learnt metrics does not significantly improve over the top proxy metric, when considered in isolation.

Nonetheless, this is not how evaluation metrics are used in practice.
Indeed, we track several metrics and can draw conclusions if \emph{any} of them are statistically significant.
As such, metrics should be evaluated on their \emph{complementary} sensitivity.
That is, we compute $p$-values for a set of metrics, apply a Bonferroni correction, and assess statistical significance.
The statistical power that we obtain through this procedure is visualised in Figure~\ref{fig:power_combined}.
We consider either the North Star in isolation, the North Star in conjunction with the top-proxy, or a further combination with any learnt metric.
Here, we observe that the learnt metric provides a substantial increase in statistical power: statistical power (i.e. $1 -$ type-II error) is increased by up to a relative 210\% compared to the North Star alone, and 25--30\% over the North Star plus proxies. 
Furthermore, as the Bonferroni correction is slightly conservative, we observe \emph{lower} than expected type-I error for higher significance levels $\alpha$.
This implies that a more fine-grained multiple testing correction can further improve statistical power.
We empirically observe that this works as expected, but its effects are negligible in practice.

\subsection{Cost Reduction from Learnt Metrics (RQ5)}
So far, we have shown that metrics learnt to minimise ($\log$)$p$-values are effective at improving sensitivity (Table~\ref{tab:objectives}), whilst minimising type-III error (Figure~\ref{fig:agreement}) and improving statistical power (Figure~\ref{fig:power}).

On one hand, powerful learnt metrics can lead to more confident decisions from statistically significant A/B-test outcomes.
Another view is that we could make the same amount of decisions based on fewer data points, as we reach statistical significance with smaller sample sizes.
This implies a cost reduction, as we can run experiments either on smaller portions of user traffic or for shorter periods of time, directly impacting experimentation velocity.

\begin{figure}[!t]
%\vspace{-3ex}
    \centering
    \includegraphics[width=\linewidth]{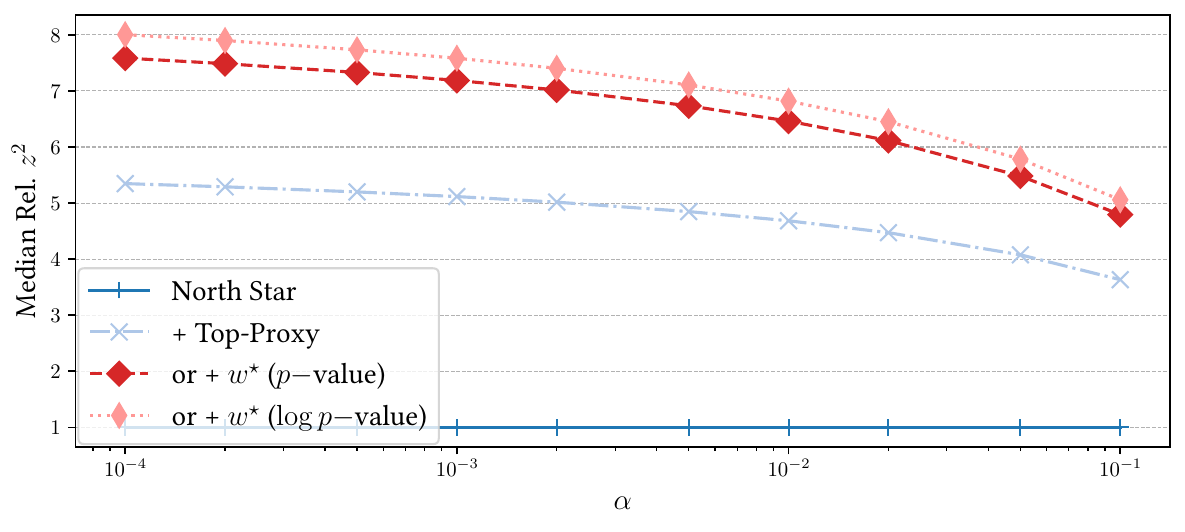}
    \caption{When considering learnt metrics in conjunction with the North Star and top proxy metrics, we require significantly reduced sample sizes to obtain the same statistical significance level as we would get from the North Star.}
    \label{fig:samplesize}
\end{figure}

This reduction in required sample size is equal to the square of the relative $z$-score~\cite{Chapelle2012,Kharitonov2017}.
We visualise this quantity in Figure~\ref{fig:samplesize}, for varying significance levels $\alpha$, for the same Bonferroni-corrected procedure as Figure~\ref{fig:power_combined}.
To obtain a $z$-score for a \emph{set} of metrics, we simply take the maximal score and apply a Bonferroni correction to it as laid out in Section~\ref{sec:pvalue_corrections}.
Note that this procedure depends on $\alpha$, explaining the slope in Figure~\ref{fig:samplesize}.
We observe that our learnt metrics can achieve the same level of statistical confidence as the North Star with up to 8 times fewer samples, i.e. a reduction down to $12.5\%$.
This significantly reduces the cost of experimentation for the business, further strengthening the case for our learnt metrics.

\begin{figure*}[t!]
    \centering
    \includegraphics[width=\linewidth]{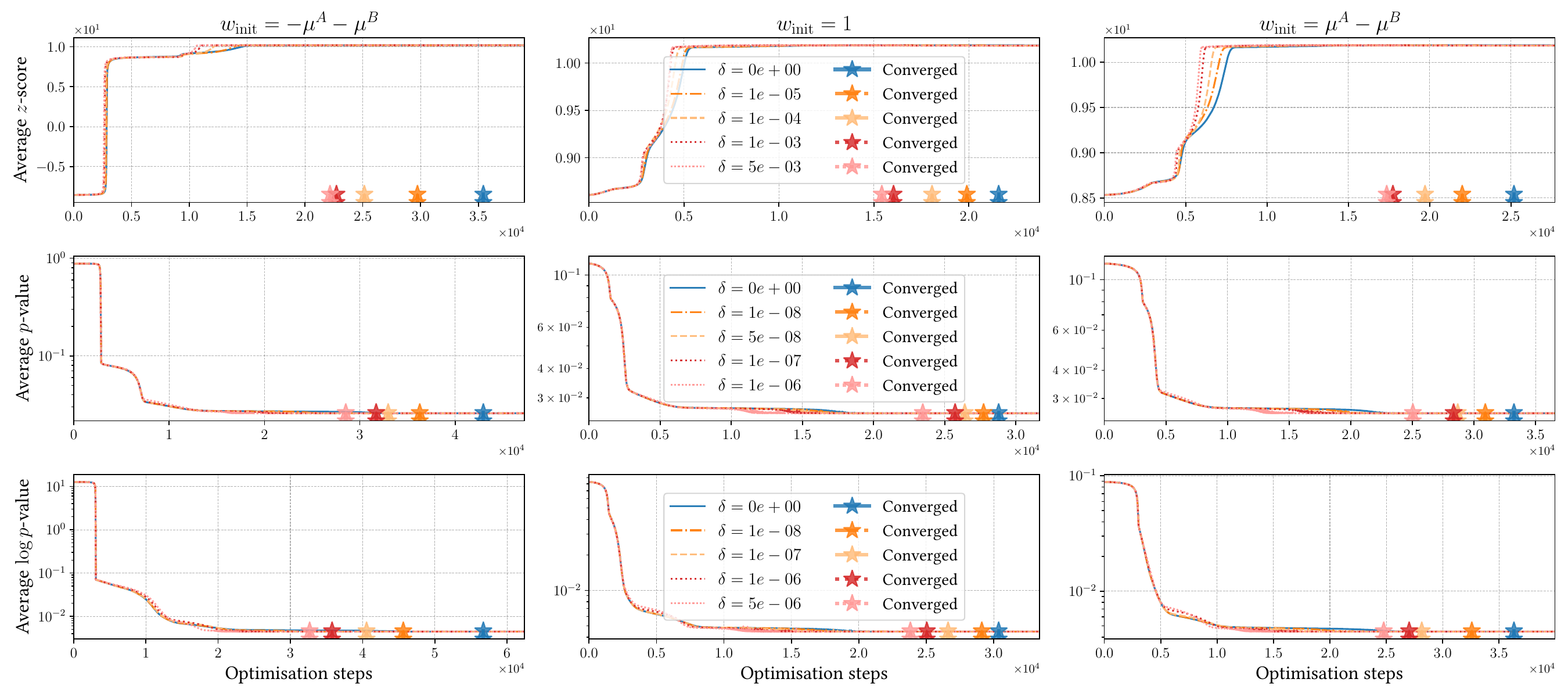}
    \caption{Spherical regularisation significantly accelerates convergence for all considered objectives, up to 40\%. }
    \label{fig:regular}
\end{figure*}

\subsection{Spherical Regularisation (RQ6)}
Our goal is to assess and quantify the effects of the proposed spherical regularisation method in Section~\ref{sec:accelerate}.
We train models on all available data with \emph{known} outcomes $\mathcal{E}^{+}$, where we have a vetted preference over variants $A \succ B$.
We consider three weight initialisation strategies to set $w_{\rm init}$, and normalise weights to ensure $\mathcal{L}_{\lVert w_{\rm init}\rVert} = 0$:
\begin{enumerate*}[label=(\roman*)]
    \item \emph{good} initialisation at $w_{\rm init}=\frac{1}{|\mathcal{E}^{+}|}\sum_{(A,B) \in \mathcal{E}^{+}}\bm{\mu}^{A}-\bm{\mu}^{B}$,
    \item \emph{constant} initialisation at the all-one vector $w_{\rm init} = \vec{\bm{1}}$, and
    \item \emph{bad} initialisation at $w_{\rm init}=\frac{1}{|\mathcal{E}^{+}|}\sum_{(A,B) \in \mathcal{E}^{+}}\bm{\mu}^{B}-\bm{\mu}^{A}$.
\end{enumerate*}
We train models for all learning objectives we deal with in this paper: $z$-scores, $p$-values, and $\log p$-values; whilst varying the strength of the spherical regularisation term $\delta$.
As discussed, this term does not affect the optima, but simply transforms the loss to be more amenable to gradient-based optimisation methods.
Thus, we expect convergence after fewer training iterations.
All models are trained until convergence with the \textsc{adam} optimiser~\cite{Kingma2014}, initialising the learning rate at $\num{5e-4}$ and halving it every 1\,000 steps where we do not observe improvements.
We use the \textsc{radam} variant to avoid convergence issues~\cite{Reddi2018,Liu2020}, and have validated that this choice does not significantly alter our obtained results and conclusions.
We consider a model converged if there are no improvements to the learning objective after 10\,000 steps.
All methods are implemented using Python3.9 and PyTorch~\cite{Paszke2019}.

Figure~\ref{fig:regular} visualises the evolution of the learning objective over optimisation steps, for all mentioned learning objectives, initialisation strategies, and regularisation strengths.
We observe that the method is robust, significantly improving convergence speed for all settings, requiring up to 40\% fewer iterations until convergence is reached.
This positively influences the practical utility of the learnt metric pipeline for researchers and practitioners.

We provide source code to reproduce Figure~\ref{fig:regularisation_effect} and our regularisation method at \href{https://github.com/olivierjeunen/learnt-metrics-kdd-2024}{github.com/olivierjeunen/learnt-metrics-kdd-2024}.

%% file: 5_Insights.tex
\section{Insights from Learnt Metrics}
In this Section, we briefly discuss insights that arose through our empirical evaluation of all metrics: the North Star, classical surrogates and proxies, as well as learnt metrics.
These insights are specific to our platforms, but we believe they can contribute to a general intuition and understanding of metrics for online content platforms and broader application areas.

\paragraph{Ratio metrics are easily fooled.}
Often, important metrics can be framed as a ratio of the means (or sums) of two existing metrics~\cite{Baweja2024,Budylin2018}.
Examples include click-through rate (i.e. clicks / impressions), variants of user retention (i.e. retained users / active users), or general engagement ratios (e.g. likes / video-plays).
We observe that, whilst these metrics can be important from a business perspective, they typically exhibit significant type-III/S errors w.r.t. the North Star.
Indeed, in the examples above both the numerator and denominator represent positive signals we wish to increase.
Suppose an online experiment increases the number of video-plays by $Y\%$, and the overall number of likes by $X\%<Y\%$.
These two positive signals will lead to a \emph{decreasing} ratio, whilst we are likely to still prefer the treatment w.r.t. the North Star if $X$ and $Y$ are substantially large.
Similar observations cautioning the use of ratio metrics have been made by \citet{Dmitriev2017}.

We believe this is connected to common \emph{offline} ranking evaluation metrics prevalent in the recommender systems field~\cite{Steck2013,Jeunen2019}.
Indeed, such metrics are cumulative in nature, optimising \emph{overall} value instead of some notion of value-per-item~\cite{Jeunen2023_nDCG}.

\paragraph{User-level aggregations conquer general counters.}
In the previous example, we describe general count metrics for the number of likes and the number of video-play events.
User behaviour on online platforms often follows a power-law distribution: a few ``power users'' generate the majority of such events~\cite{Chi2020}.
As a result, such metrics are easily skewed, and they are not guaranteed to accurately reflect improvements for the full population of users---empirically leading to type-III/S errors w.r.t. the North Star.
Aggregating such counters per users (e.g. count the number of days a user has at least $X$ video-plays) instead of using raw event counters, provides strong and sensitive proxies to the North Star.

Interestingly, this framing is reminiscent of \emph{recall}, as we effectively measure the coverage of users about whom we have positive signals.
Recall metrics are again strongly connected to offline evaluation practices in recommender systems, especially in the first stage of two-stage systems common in industry~\cite{Covington2016,ma2020off}.

%% file: 6_Conclusions.tex
\section{Conclusions \& Outlook}
A/B-testing is a crucial tool for decision-making in online businesses, and it has widely been adopted as a go-to approach to allow for continuous system improvements.
Notwithstanding their popularity, online experiments are often expensive to perform.
Indeed: many experiments lead to statistically insignificant outcomes, presenting an obstacle for confident decision-making.
Experiments that do lead to significant outcomes are costly too: by their very definition, a portion of user traffic interacts with a sub-optimal system variant.
As such, we want to maximise the number of decisions we can make based on the experiments we run, and we want to minimise the required sample size for statistically significant outcomes.
In this work, we achieve this by learning metrics that maximise the statistical power they harness.
We present novel learning objectives for such metrics, and provide a thorough evaluation of the effectiveness of our proposed approaches.
Our learnt metrics are currently used for confident, high-velocity decision-making across ShareChat and Moj business units.

We believe our work opens several avenues for future work improving the efficacy of learnt metrics, by e.g. relaxing the linearity constraint we rely on.
Furthermore, we wish to leverage our learnt metrics as reward signals for personalisation through machine learning models~\cite{Jeunen2021Thesis}.

%% file: A_Reproducibility.tex
%\newpage
\begin{figure*}[h]
    %\vspace{-3ex}
    \begin{subfigure}[t]{\textwidth}
        \centering
        \includegraphics[width=\linewidth]{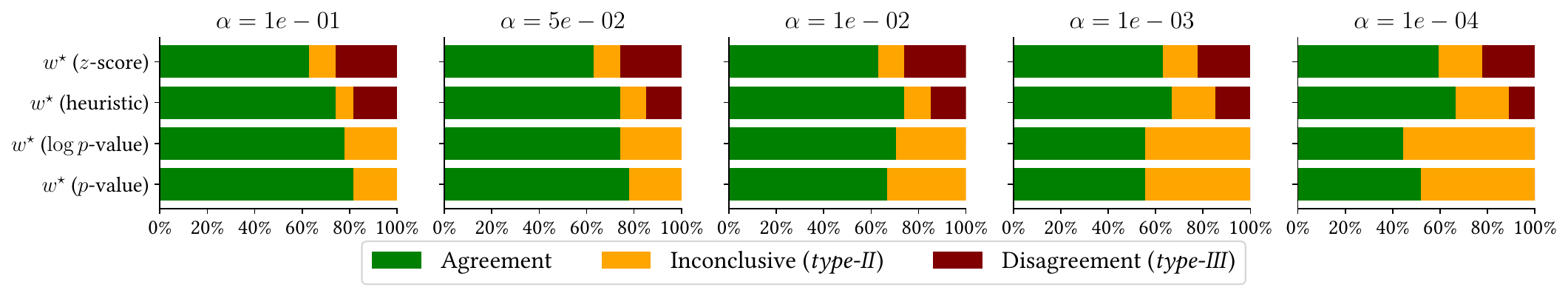}
    \caption{Figure~\ref{fig:agreement} reproduced with data from Moj.}\label{fig:agreement_Moj}
    \end{subfigure}
    ~\\
    \begin{subfigure}[t]{\textwidth}
        \centering
        \includegraphics[width=\linewidth]{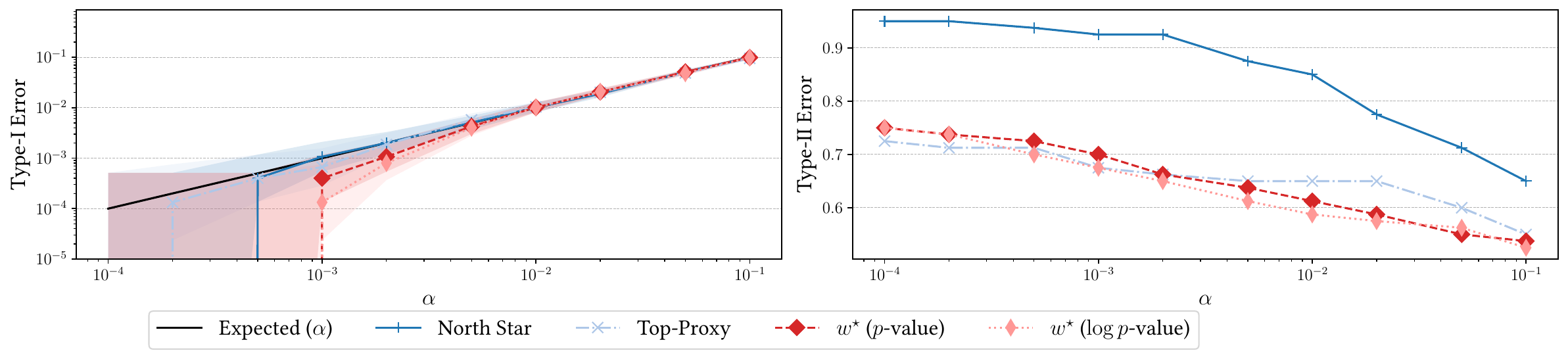}
    \caption{Figure~\ref{fig:power_standalone} reproduced with data from Moj.}\label{fig:power_standalone_Moj}
    \end{subfigure}
    ~\\
    \begin{subfigure}[t]{\textwidth}
        \centering
        \includegraphics[width=\linewidth]{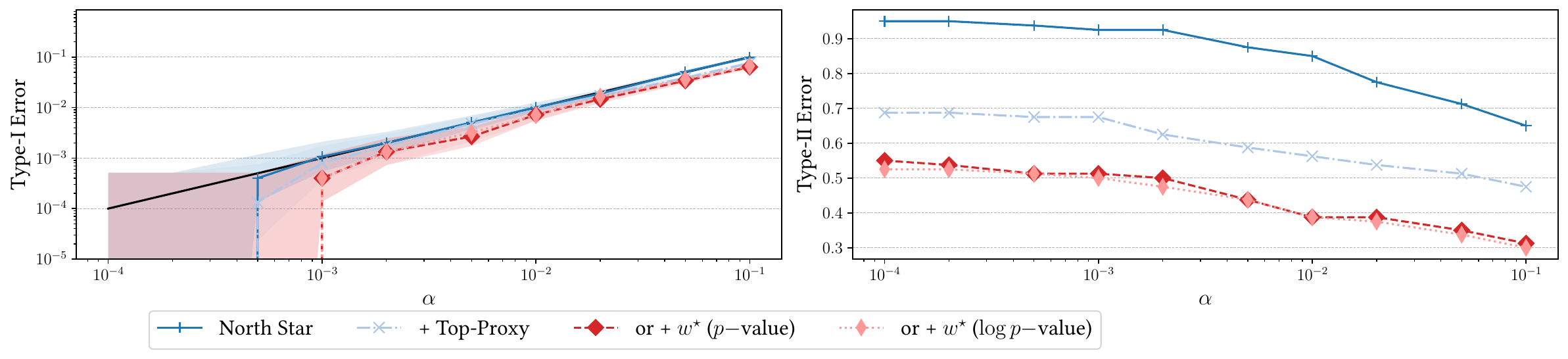}
    \caption{Figure~\ref{fig:power_combined} reproduced with data from Moj.}\label{fig:power_combined_Moj}
    \end{subfigure}
    \begin{subfigure}[t]{0.5\textwidth}
        \centering
        \includegraphics[width=\linewidth]{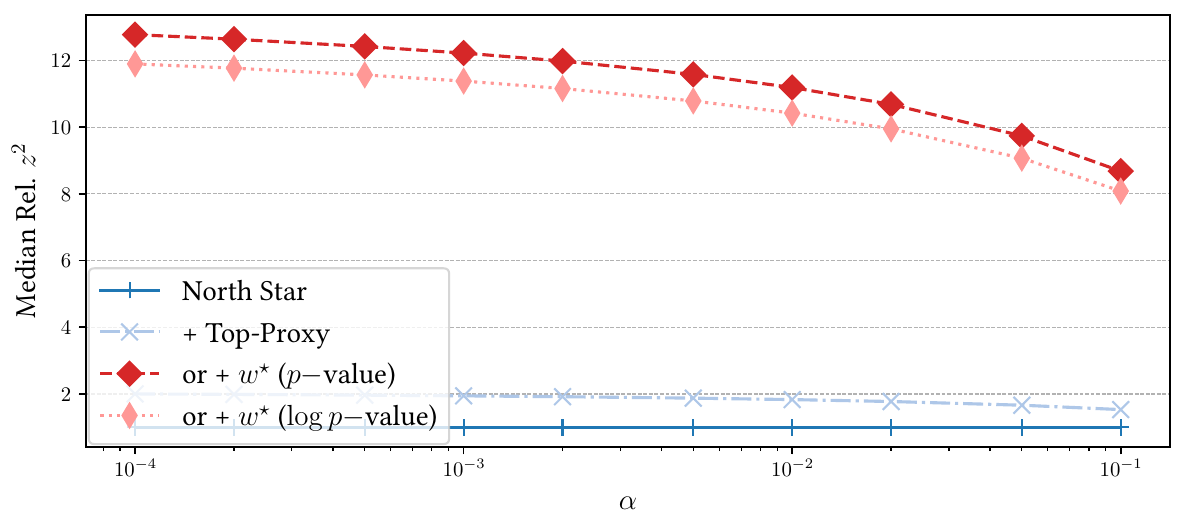}
    \caption{Figure~\ref{fig:samplesize} reproduced with data from Moj. }\label{fig:samplesize_Moj}
    \end{subfigure}
    \caption{Additional experimental results for Moj. Empirical observations match those for ShareChat.}\label{fig:power_Moj}
\end{figure*}
\newpage
\section{Additional Experimental Results}\label{sec:appendix}
To further empirically validate our theoretical insights w.r.t. the proposed methods, we repeat the experiments reported in Section~\ref{sec:experiments} on data collected for the Moj platform, and reproduce Figures~\ref{fig:agreement}--\ref{fig:samplesize}.
Results are visualised in Figure~\ref{fig:power_Moj}.
Observations match our expectations, further strengthening trust in the replicability of our results.

All improvements in sensitivity and statistical power are a similar order of magnitude as those for ShareChat: learnt metrics that minimise ($\log$)$p$-values can substantially reduce type-II/III errors without affecting type-I errors.
We observe an improvement over ShareChat data in Figure~\ref{fig:samplesize_Moj}: learnt metrics for Moj exhibit a 12-fold reduction in the sample size that is required to attain constant statistical confidence as to the North Star.

%% file: main.bbl

\begin{thebibliography}{44}

%%% the Web address.

\ifx \showCODEN    \undefined \def \showCODEN     #1{\unskip}     \fi
\ifx \showDOI      \undefined \def \showDOI       #1{#1}\fi
\ifx \showISBNx    \undefined \def \showISBNx     #1{\unskip}     \fi
\ifx \showISBNxiii \undefined \def \showISBNxiii  #1{\unskip}     \fi
\ifx \showISSN     \undefined \def \showISSN      #1{\unskip}     \fi
\ifx \showLCCN     \undefined \def \showLCCN      #1{\unskip}     \fi
\ifx \shownote     \undefined \def \shownote      #1{#1}          \fi
\ifx \showarticletitle \undefined \def \showarticletitle #1{#1}   \fi
\ifx \showURL      \undefined \def \showURL       {\relax}        \fi
% invisible to TeX
\providecommand\bibfield[2]{#2}
\providecommand\bibinfo[2]{#2}
\providecommand\natexlab[1]{#1}
\providecommand\showeprint[2][]{arXiv:#2}

\bibitem[Athey et~al\mbox{.}(2019)]%
        {Athey2019}
\bibfield{author}{\bibinfo{person}{Susan Athey}, \bibinfo{person}{Raj Chetty}, \bibinfo{person}{Guido~W Imbens}, {and} \bibinfo{person}{Hyunseung Kang}.} \bibinfo{year}{2019}\natexlab{}.
\newblock \bibinfo{booktitle}{\emph{The Surrogate Index: Combining Short-Term Proxies to Estimate Long-Term Treatment Effects More Rapidly and Precisely}}.
\newblock \bibinfo{type}{Working Paper} 26463. \bibinfo{institution}{National Bureau of Economic Research}.
\newblock
\urldef\tempurl%
\url{https://doi.org/10.3386/w26463}
\showDOI{\tempurl}


\bibitem[Baweja et~al\mbox{.}(2024)]%
        {Baweja2024}
\bibfield{author}{\bibinfo{person}{Shubham Baweja}, \bibinfo{person}{Neeti Pokharna}, \bibinfo{person}{Aleksei Ustimenko}, {and} \bibinfo{person}{Olivier Jeunen}.} \bibinfo{year}{2024}\natexlab{}.
\newblock \showarticletitle{Variance Reduction in Ratio Metrics for Efficient Online Experiments}. In \bibinfo{booktitle}{\emph{Proc. of the 46th European Conference on Information Retrieval}} \emph{(\bibinfo{series}{ECIR '24})}. \bibinfo{publisher}{Springer}.
\newblock


\bibitem[Budylin et~al\mbox{.}(2018)]%
        {Budylin2018}
\bibfield{author}{\bibinfo{person}{Roman Budylin}, \bibinfo{person}{Alexey Drutsa}, \bibinfo{person}{Ilya Katsev}, {and} \bibinfo{person}{Valeriya Tsoy}.} \bibinfo{year}{2018}\natexlab{}.
\newblock \showarticletitle{Consistent Transformation of Ratio Metrics for Efficient Online Controlled Experiments}. In \bibinfo{booktitle}{\emph{Proc. of the Eleventh ACM International Conference on Web Search and Data Mining}} \emph{(\bibinfo{series}{WSDM '18})}. \bibinfo{publisher}{ACM}, \bibinfo{pages}{55–63}.
\newblock
\showISBNx{9781450355810}
\urldef\tempurl%
\url{https://doi.org/10.1145/3159652.3159699}
\showDOI{\tempurl}


\bibitem[Chapelle et~al\mbox{.}(2012)]%
        {Chapelle2012}
\bibfield{author}{\bibinfo{person}{Olivier Chapelle}, \bibinfo{person}{Thorsten Joachims}, \bibinfo{person}{Filip Radlinski}, {and} \bibinfo{person}{Yisong Yue}.} \bibinfo{year}{2012}\natexlab{}.
\newblock \showarticletitle{Large-Scale Validation and Analysis of Interleaved Search Evaluation}.
\newblock \bibinfo{journal}{\emph{ACM Trans. Inf. Syst.}} \bibinfo{volume}{30}, \bibinfo{number}{1}, Article \bibinfo{articleno}{6} (\bibinfo{date}{mar} \bibinfo{year}{2012}), \bibinfo{numpages}{41}~pages.
\newblock
\showISSN{1046-8188}
\urldef\tempurl%
\url{https://doi.org/10.1145/2094072.2094078}
\showDOI{\tempurl}


\bibitem[Chi(2020)]%
        {Chi2020}
\bibfield{author}{\bibinfo{person}{Ed~H. Chi}.} \bibinfo{year}{2020}\natexlab{}.
\newblock \showarticletitle{From Missing Data to Boltzmann Distributions and Time Dynamics: The Statistical Physics of Recommendation}. In \bibinfo{booktitle}{\emph{Proc. of the 13th International Conference on Web Search and Data Mining}} \emph{(\bibinfo{series}{WSDM '20})}. \bibinfo{publisher}{ACM}, \bibinfo{pages}{1–2}.
\newblock
\showISBNx{9781450368223}
\urldef\tempurl%
\url{https://doi.org/10.1145/3336191.3372193}
\showDOI{\tempurl}


\bibitem[Covington et~al\mbox{.}(2016)]%
        {Covington2016}
\bibfield{author}{\bibinfo{person}{Paul Covington}, \bibinfo{person}{Jay Adams}, {and} \bibinfo{person}{Emre Sargin}.} \bibinfo{year}{2016}\natexlab{}.
\newblock \showarticletitle{Deep Neural Networks for YouTube Recommendations}. In \bibinfo{booktitle}{\emph{Proc. of the 10th ACM Conference on Recommender Systems}} \emph{(\bibinfo{series}{RecSys '16})}. \bibinfo{publisher}{ACM}, \bibinfo{pages}{191–198}.
\newblock
\showISBNx{9781450340359}
\urldef\tempurl%
\url{https://doi.org/10.1145/2959100.2959190}
\showDOI{\tempurl}


\bibitem[Deng and Shi(2016)]%
        {Deng2016}
\bibfield{author}{\bibinfo{person}{Alex Deng} {and} \bibinfo{person}{Xiaolin Shi}.} \bibinfo{year}{2016}\natexlab{}.
\newblock \showarticletitle{Data-Driven Metric Development for Online Controlled Experiments: Seven Lessons Learned}. In \bibinfo{booktitle}{\emph{Proc. of the 22nd ACM SIGKDD International Conference on Knowledge Discovery and Data Mining}} \emph{(\bibinfo{series}{KDD '16})}. \bibinfo{publisher}{ACM}, \bibinfo{pages}{77–86}.
\newblock
\showISBNx{9781450342322}
\urldef\tempurl%
\url{https://doi.org/10.1145/2939672.2939700}
\showDOI{\tempurl}


\bibitem[Deng et~al\mbox{.}(2013)]%
        {Deng2013}
\bibfield{author}{\bibinfo{person}{Alex Deng}, \bibinfo{person}{Ya Xu}, \bibinfo{person}{Ron Kohavi}, {and} \bibinfo{person}{Toby Walker}.} \bibinfo{year}{2013}\natexlab{}.
\newblock \showarticletitle{Improving the Sensitivity of Online Controlled Experiments by Utilizing Pre-Experiment Data}. In \bibinfo{booktitle}{\emph{Proc. of the Sixth ACM International Conference on Web Search and Data Mining}} \emph{(\bibinfo{series}{WSDM '13})}. \bibinfo{publisher}{ACM}, \bibinfo{pages}{123–132}.
\newblock
\showISBNx{9781450318693}
\urldef\tempurl%
\url{https://doi.org/10.1145/2433396.2433413}
\showDOI{\tempurl}


\bibitem[Dmitriev et~al\mbox{.}(2017)]%
        {Dmitriev2017}
\bibfield{author}{\bibinfo{person}{Pavel Dmitriev}, \bibinfo{person}{Somit Gupta}, \bibinfo{person}{Dong~Woo Kim}, {and} \bibinfo{person}{Garnet Vaz}.} \bibinfo{year}{2017}\natexlab{}.
\newblock \showarticletitle{A Dirty Dozen: Twelve Common Metric Interpretation Pitfalls in Online Controlled Experiments}. In \bibinfo{booktitle}{\emph{Proceedings of the 23rd ACM SIGKDD International Conference on Knowledge Discovery and Data Mining}} \emph{(\bibinfo{series}{KDD '17})}. \bibinfo{publisher}{ACM}, \bibinfo{pages}{1427–1436}.
\newblock
\showISBNx{9781450348874}
\urldef\tempurl%
\url{https://doi.org/10.1145/3097983.3098024}
\showDOI{\tempurl}


\bibitem[Gelman and Carlin(2014)]%
        {Gelman2014}
\bibfield{author}{\bibinfo{person}{Andrew Gelman} {and} \bibinfo{person}{John Carlin}.} \bibinfo{year}{2014}\natexlab{}.
\newblock \showarticletitle{Beyond Power Calculations: Assessing Type S (Sign) and Type M (Magnitude) Errors}.
\newblock \bibinfo{journal}{\emph{Perspectives on Psychological Science}} \bibinfo{volume}{9}, \bibinfo{number}{6} (\bibinfo{year}{2014}), \bibinfo{pages}{641--651}.
\newblock
\urldef\tempurl%
\url{https://doi.org/10.1177/1745691614551642}
\showDOI{\tempurl}
\newblock
\shownote{PMID: 26186114}.


\bibitem[Goffrier et~al\mbox{.}(2023)]%
        {Goffrier2023}
\bibfield{author}{\bibinfo{person}{Graham~Van Goffrier}, \bibinfo{person}{Lucas Maystre}, {and} \bibinfo{person}{Ciar\'an~Mark Gilligan-Lee}.} \bibinfo{year}{2023}\natexlab{}.
\newblock \showarticletitle{Estimating long-term causal effects from short-term experiments and long-term observational data with unobserved confounding}. In \bibinfo{booktitle}{\emph{Proc. of the Second Conference on Causal Learning and Reasoning}} \emph{(\bibinfo{series}{Proc. of Machine Learning Research}, Vol.~\bibinfo{volume}{213})}, \bibfield{editor}{\bibinfo{person}{Mihaela van~der Schaar}, \bibinfo{person}{Cheng Zhang}, {and} \bibinfo{person}{Dominik Janzing}} (Eds.). \bibinfo{publisher}{PMLR}, \bibinfo{pages}{791--813}.
\newblock
\urldef\tempurl%
\url{https://proceedings.mlr.press/v213/goffrier23a.html}
\showURL{%
\tempurl}


\bibitem[Guo et~al\mbox{.}(2021)]%
        {Guo2021}
\bibfield{author}{\bibinfo{person}{Yongyi Guo}, \bibinfo{person}{Dominic Coey}, \bibinfo{person}{Mikael Konutgan}, \bibinfo{person}{Wenting Li}, \bibinfo{person}{Chris Schoener}, {and} \bibinfo{person}{Matt Goldman}.} \bibinfo{year}{2021}\natexlab{}.
\newblock \showarticletitle{Machine Learning for Variance Reduction in Online Experiments}. In \bibinfo{booktitle}{\emph{Advances in Neural Information Processing Systems}}, Vol.~\bibinfo{volume}{34}. \bibinfo{publisher}{Curran Associates, Inc.}, \bibinfo{pages}{8637--8648}.
\newblock


\bibitem[Howard et~al\mbox{.}(2021)]%
        {Howard2021}
\bibfield{author}{\bibinfo{person}{Steven~R. Howard}, \bibinfo{person}{Aaditya Ramdas}, \bibinfo{person}{Jon McAuliffe}, {and} \bibinfo{person}{Jasjeet Sekhon}.} \bibinfo{year}{2021}\natexlab{}.
\newblock \showarticletitle{{Time-uniform, nonparametric, nonasymptotic confidence sequences}}.
\newblock \bibinfo{journal}{\emph{The Annals of Statistics}} \bibinfo{volume}{49}, \bibinfo{number}{2} (\bibinfo{year}{2021}), \bibinfo{pages}{1055 -- 1080}.
\newblock
\urldef\tempurl%
\url{https://doi.org/10.1214/20-AOS1991}
\showDOI{\tempurl}


\bibitem[Jeunen(2019)]%
        {Jeunen2019}
\bibfield{author}{\bibinfo{person}{Olivier Jeunen}.} \bibinfo{year}{2019}\natexlab{}.
\newblock \showarticletitle{Revisiting Offline Evaluation for Implicit-Feedback Recommender Systems}. In \bibinfo{booktitle}{\emph{Proc. of the 13th ACM Conference on Recommender Systems}} \emph{(\bibinfo{series}{RecSys '19})}. \bibinfo{publisher}{ACM}, \bibinfo{pages}{596–600}.
\newblock
\showISBNx{9781450362436}
\urldef\tempurl%
\url{https://doi.org/10.1145/3298689.3347069}
\showDOI{\tempurl}


\bibitem[Jeunen(2021)]%
        {Jeunen2021Thesis}
\bibfield{author}{\bibinfo{person}{Olivier Jeunen}.} \bibinfo{year}{2021}\natexlab{}.
\newblock \emph{\bibinfo{title}{Offline Approaches to Recommendation with Online Success}}.
\newblock \bibinfo{thesistype}{Ph.\,D. Dissertation}. \bibinfo{school}{University of Antwerp}.
\newblock


\bibitem[Jeunen(2023)]%
        {Jeunen2023_Forum}
\bibfield{author}{\bibinfo{person}{Olivier Jeunen}.} \bibinfo{year}{2023}\natexlab{}.
\newblock \showarticletitle{A Common Misassumption in Online Experiments with Machine Learning Models}.
\newblock \bibinfo{journal}{\emph{SIGIR Forum}} \bibinfo{volume}{57}, \bibinfo{number}{1}, Article \bibinfo{articleno}{13} (\bibinfo{date}{dec} \bibinfo{year}{2023}), \bibinfo{numpages}{9}~pages.
\newblock
\showISSN{0163-5840}
\urldef\tempurl%
\url{https://doi.org/10.1145/3636341.3636358}
\showDOI{\tempurl}


\bibitem[Jeunen et~al\mbox{.}(2024)]%
        {Jeunen2023_nDCG}
\bibfield{author}{\bibinfo{person}{Olivier Jeunen}, \bibinfo{person}{Ivan Potapov}, {and} \bibinfo{person}{Aleksei Ustimenko}.} \bibinfo{year}{2024}\natexlab{}.
\newblock \showarticletitle{On (Normalised) Discounted Cumulative Gain as an Offline Evaluation Metric for Top-$n$ Recommendation}. In \bibinfo{booktitle}{\emph{Proc. of the 26th ACM SIGKDD International Conference on Knowledge Discovery \& Data Mining}} \emph{(\bibinfo{series}{KDD '24})}.
\newblock
\showeprint[arxiv]{2307.15053}~[cs.IR]


\bibitem[Kaiser(1960)]%
        {Kaiser1960}
\bibfield{author}{\bibinfo{person}{Henry~F Kaiser}.} \bibinfo{year}{1960}\natexlab{}.
\newblock \showarticletitle{Directional statistical decisions.}
\newblock \bibinfo{journal}{\emph{Psychological Review}} \bibinfo{volume}{67}, \bibinfo{number}{3} (\bibinfo{year}{1960}), \bibinfo{pages}{160}.
\newblock


\bibitem[Kharitonov et~al\mbox{.}(2017)]%
        {Kharitonov2017}
\bibfield{author}{\bibinfo{person}{Eugene Kharitonov}, \bibinfo{person}{Alexey Drutsa}, {and} \bibinfo{person}{Pavel Serdyukov}.} \bibinfo{year}{2017}\natexlab{}.
\newblock \showarticletitle{Learning Sensitive Combinations of A/B Test Metrics}. In \bibinfo{booktitle}{\emph{Proc. of the Tenth ACM International Conference on Web Search and Data Mining}} \emph{(\bibinfo{series}{WSDM '17})}. \bibinfo{publisher}{ACM}, \bibinfo{pages}{651–659}.
\newblock
\showISBNx{9781450346757}
\urldef\tempurl%
\url{https://doi.org/10.1145/3018661.3018708}
\showDOI{\tempurl}


\bibitem[Kingma and Ba(2014)]%
        {Kingma2014}
\bibfield{author}{\bibinfo{person}{Diederik~P. Kingma} {and} \bibinfo{person}{Jimmy Ba}.} \bibinfo{year}{2014}\natexlab{}.
\newblock \showarticletitle{Adam: A Method for Stochastic Optimization}. In \bibinfo{booktitle}{\emph{Proc. of the 3rd International Conference on Learning Representations}} \emph{(\bibinfo{series}{ICLR '14})}.
\newblock
\showeprint[arxiv]{1412.6980}~[cs.LG]


\bibitem[Kohavi et~al\mbox{.}(2022)]%
        {Kohavi2022}
\bibfield{author}{\bibinfo{person}{Ron Kohavi}, \bibinfo{person}{Alex Deng}, {and} \bibinfo{person}{Lukas Vermeer}.} \bibinfo{year}{2022}\natexlab{}.
\newblock \showarticletitle{A/B Testing Intuition Busters: Common Misunderstandings in Online Controlled Experiments}. In \bibinfo{booktitle}{\emph{Proc. of the 28th ACM SIGKDD Conference on Knowledge Discovery and Data Mining}} \emph{(\bibinfo{series}{KDD '22})}. \bibinfo{publisher}{ACM}, \bibinfo{pages}{3168–3177}.
\newblock
\showISBNx{9781450393850}
\urldef\tempurl%
\url{https://doi.org/10.1145/3534678.3539160}
\showURL{%
\tempurl}


\bibitem[Kohavi et~al\mbox{.}(2020)]%
        {kohavi2020trustworthy}
\bibfield{author}{\bibinfo{person}{Ron Kohavi}, \bibinfo{person}{Diane Tang}, {and} \bibinfo{person}{Ya Xu}.} \bibinfo{year}{2020}\natexlab{}.
\newblock \bibinfo{booktitle}{\emph{Trustworthy online controlled experiments: A practical guide to A/B testing}}.
\newblock \bibinfo{publisher}{Cambridge University Press}.
\newblock


\bibitem[Ledoit and Wolf(2004)]%
        {Ledoit2004}
\bibfield{author}{\bibinfo{person}{Olivier Ledoit} {and} \bibinfo{person}{Michael Wolf}.} \bibinfo{year}{2004}\natexlab{}.
\newblock \showarticletitle{A well-conditioned estimator for large-dimensional covariance matrices}.
\newblock \bibinfo{journal}{\emph{Journal of Multivariate Analysis}} \bibinfo{volume}{88}, \bibinfo{number}{2} (\bibinfo{year}{2004}), \bibinfo{pages}{365--411}.
\newblock
\showISSN{0047-259X}
\urldef\tempurl%
\url{https://doi.org/10.1016/S0047-259X(03)00096-4}
\showDOI{\tempurl}


\bibitem[Ledoit and Wolf(2020)]%
        {Ledoit2020}
\bibfield{author}{\bibinfo{person}{Olivier Ledoit} {and} \bibinfo{person}{Michael Wolf}.} \bibinfo{year}{2020}\natexlab{}.
\newblock \showarticletitle{{The Power of (Non-)Linear Shrinking: A Review and Guide to Covariance Matrix Estimation}}.
\newblock \bibinfo{journal}{\emph{Journal of Financial Econometrics}} \bibinfo{volume}{20}, \bibinfo{number}{1} (\bibinfo{date}{06} \bibinfo{year}{2020}), \bibinfo{pages}{187--218}.
\newblock
\showISSN{1479-8409}
\urldef\tempurl%
\url{https://doi.org/10.1093/jjfinec/nbaa007}
\showDOI{\tempurl}
\showeprint{https://academic.oup.com/jfec/article-pdf/20/1/187/42274902/nbaa007.pdf}


\bibitem[Liu et~al\mbox{.}(2020)]%
        {Liu2020}
\bibfield{author}{\bibinfo{person}{Liyuan Liu}, \bibinfo{person}{Haoming Jiang}, \bibinfo{person}{Pengcheng He}, \bibinfo{person}{Weizhu Chen}, \bibinfo{person}{Xiaodong Liu}, \bibinfo{person}{Jianfeng Gao}, {and} \bibinfo{person}{Jiawei Han}.} \bibinfo{year}{2020}\natexlab{}.
\newblock \showarticletitle{On the Variance of the Adaptive Learning Rate and Beyond}. In \bibinfo{booktitle}{\emph{International Conference on Learning Representations}} \emph{(\bibinfo{series}{ICLR '20})}.
\newblock
\urldef\tempurl%
\url{https://arxiv.org/abs/1908.03265}
\showURL{%
\tempurl}


\bibitem[Ma et~al\mbox{.}(2020)]%
        {ma2020off}
\bibfield{author}{\bibinfo{person}{Jiaqi Ma}, \bibinfo{person}{Zhe Zhao}, \bibinfo{person}{Xinyang Yi}, \bibinfo{person}{Ji Yang}, \bibinfo{person}{Minmin Chen}, \bibinfo{person}{Jiaxi Tang}, \bibinfo{person}{Lichan Hong}, {and} \bibinfo{person}{Ed~H Chi}.} \bibinfo{year}{2020}\natexlab{}.
\newblock \showarticletitle{Off-policy learning in two-stage recommender systems}. In \bibinfo{booktitle}{\emph{Proc. of The Web Conference 2020}}. \bibinfo{pages}{463--473}.
\newblock


\bibitem[Mosteller(1948)]%
        {Mosteller1948}
\bibfield{author}{\bibinfo{person}{Frederick Mosteller}.} \bibinfo{year}{1948}\natexlab{}.
\newblock \showarticletitle{A k-Sample Slippage Test for an Extreme Population}.
\newblock \bibinfo{journal}{\emph{The Annals of Mathematical Statistics}} \bibinfo{volume}{19}, \bibinfo{number}{1} (\bibinfo{year}{1948}), \bibinfo{pages}{58--65}.
\newblock
\showISSN{00034851}
\urldef\tempurl%
\url{http://www.jstor.org/stable/2236056}
\showURL{%
\tempurl}


\bibitem[Paszke et~al\mbox{.}(2019)]%
        {Paszke2019}
\bibfield{author}{\bibinfo{person}{Adam Paszke}, \bibinfo{person}{Sam Gross}, \bibinfo{person}{Francisco Massa}, \bibinfo{person}{Adam Lerer}, \bibinfo{person}{James Bradbury}, \bibinfo{person}{Gregory Chanan}, \bibinfo{person}{Trevor Killeen}, \bibinfo{person}{Zeming Lin}, \bibinfo{person}{Natalia Gimelshein}, \bibinfo{person}{Luca Antiga}, \bibinfo{person}{Alban Desmaison}, \bibinfo{person}{Andreas Kopf}, \bibinfo{person}{Edward Yang}, \bibinfo{person}{Zachary DeVito}, \bibinfo{person}{Martin Raison}, \bibinfo{person}{Alykhan Tejani}, \bibinfo{person}{Sasank Chilamkurthy}, \bibinfo{person}{Benoit Steiner}, \bibinfo{person}{Lu Fang}, \bibinfo{person}{Junjie Bai}, {and} \bibinfo{person}{Soumith Chintala}.} \bibinfo{year}{2019}\natexlab{}.
\newblock \showarticletitle{PyTorch: An Imperative Style, High-Performance Deep Learning Library}. In \bibinfo{booktitle}{\emph{Advances in Neural Information Processing Systems}}, \bibfield{editor}{\bibinfo{person}{H.~Wallach}, \bibinfo{person}{H.~Larochelle}, \bibinfo{person}{A.~Beygelzimer}, \bibinfo{person}{F.~d\textquotesingle Alch\'{e}-Buc}, \bibinfo{person}{E.~Fox}, {and} \bibinfo{person}{R.~Garnett}} (Eds.), Vol.~\bibinfo{volume}{32}. \bibinfo{publisher}{Curran Associates, Inc.}
\newblock
\urldef\tempurl%
\url{https://proceedings.neurips.cc/paper_files/paper/2019/file/bdbca288fee7f92f2bfa9f7012727740-Paper.pdf}
\showURL{%
\tempurl}


\bibitem[Poyarkov et~al\mbox{.}(2016)]%
        {Poyarkov2016}
\bibfield{author}{\bibinfo{person}{Alexey Poyarkov}, \bibinfo{person}{Alexey Drutsa}, \bibinfo{person}{Andrey Khalyavin}, \bibinfo{person}{Gleb Gusev}, {and} \bibinfo{person}{Pavel Serdyukov}.} \bibinfo{year}{2016}\natexlab{}.
\newblock \showarticletitle{Boosted Decision Tree Regression Adjustment for Variance Reduction in Online Controlled Experiments}. In \bibinfo{booktitle}{\emph{Proc. of the 22nd ACM SIGKDD International Conference on Knowledge Discovery and Data Mining}} \emph{(\bibinfo{series}{KDD '16})}. \bibinfo{publisher}{ACM}, \bibinfo{pages}{235–244}.
\newblock
\showISBNx{9781450342322}
\urldef\tempurl%
\url{https://doi.org/10.1145/2939672.2939688}
\showDOI{\tempurl}


\bibitem[Reddi et~al\mbox{.}(2018)]%
        {Reddi2018}
\bibfield{author}{\bibinfo{person}{Sashank~J. Reddi}, \bibinfo{person}{Satyen Kale}, {and} \bibinfo{person}{Sanjiv Kumar}.} \bibinfo{year}{2018}\natexlab{}.
\newblock \showarticletitle{On the Convergence of Adam and Beyond}. In \bibinfo{booktitle}{\emph{International Conference on Learning Representations}} \emph{(\bibinfo{series}{ICLR '18})}.
\newblock
\urldef\tempurl%
\url{https://openreview.net/forum?id=ryQu7f-RZ}
\showURL{%
\tempurl}


\bibitem[Richardson et~al\mbox{.}(2023)]%
        {Richardson2023}
\bibfield{author}{\bibinfo{person}{Lee Richardson}, \bibinfo{person}{Alessandro Zito}, \bibinfo{person}{Dylan Greaves}, {and} \bibinfo{person}{Jacopo Soriano}.} \bibinfo{year}{2023}\natexlab{}.
\newblock \bibinfo{title}{Pareto optimal proxy metrics}.
\newblock
\newblock
\showeprint[arxiv]{2307.01000}~[stat.ME]


\bibitem[Rubin(1974)]%
        {Rubin1974}
\bibfield{author}{\bibinfo{person}{Donald~B Rubin}.} \bibinfo{year}{1974}\natexlab{}.
\newblock \showarticletitle{Estimating causal effects of treatments in randomized and nonrandomized studies.}
\newblock \bibinfo{journal}{\emph{Journal of educational Psychology}} \bibinfo{volume}{66}, \bibinfo{number}{5} (\bibinfo{year}{1974}), \bibinfo{pages}{688}.
\newblock


\bibitem[Schmit and Miller(2022)]%
        {schmit2022sequential}
\bibfield{author}{\bibinfo{person}{Sven Schmit} {and} \bibinfo{person}{Evan Miller}.} \bibinfo{year}{2022}\natexlab{}.
\newblock \showarticletitle{Sequential confidence intervals for relative lift with regression adjustments}.
\newblock  (\bibinfo{year}{2022}).
\newblock


\bibitem[Shaffer(1995)]%
        {Shaffer1995}
\bibfield{author}{\bibinfo{person}{Juliet~Popper Shaffer}.} \bibinfo{year}{1995}\natexlab{}.
\newblock \showarticletitle{Multiple Hypothesis Testing}.
\newblock \bibinfo{journal}{\emph{Annual Review of Psychology}} \bibinfo{volume}{46}, \bibinfo{number}{1} (\bibinfo{year}{1995}), \bibinfo{pages}{561--584}.
\newblock
\urldef\tempurl%
\url{https://doi.org/10.1146/annurev.ps.46.020195.003021}
\showDOI{\tempurl}
\showeprint{https://doi.org/10.1146/annurev.ps.46.020195.003021}


\bibitem[Steck(2013)]%
        {Steck2013}
\bibfield{author}{\bibinfo{person}{Harald Steck}.} \bibinfo{year}{2013}\natexlab{}.
\newblock \showarticletitle{Evaluation of recommendations: rating-prediction and ranking}. In \bibinfo{booktitle}{\emph{Proc. of the 7th ACM Conference on Recommender Systems}} \emph{(\bibinfo{series}{RecSys '13})}. \bibinfo{publisher}{ACM}, \bibinfo{pages}{213–220}.
\newblock
\showISBNx{9781450324090}
\urldef\tempurl%
\url{https://doi.org/10.1145/2507157.2507160}
\showDOI{\tempurl}


\bibitem[Tang et~al\mbox{.}(2022)]%
        {Tang2022}
\bibfield{author}{\bibinfo{person}{Ziyang Tang}, \bibinfo{person}{Yiheng Duan}, \bibinfo{person}{Steven Zhu}, \bibinfo{person}{Stephanie Zhang}, {and} \bibinfo{person}{Lihong Li}.} \bibinfo{year}{2022}\natexlab{}.
\newblock \showarticletitle{Estimating Long-Term Effects from Experimental Data}. In \bibinfo{booktitle}{\emph{Proc. of the 16th ACM Conference on Recommender Systems}} \emph{(\bibinfo{series}{RecSys '22})}. \bibinfo{publisher}{ACM}, \bibinfo{pages}{516–518}.
\newblock
\showISBNx{9781450392785}
\urldef\tempurl%
\url{https://doi.org/10.1145/3523227.3547398}
\showDOI{\tempurl}


\bibitem[Tripuraneni et~al\mbox{.}(2023)]%
        {Tripuraneni2023}
\bibfield{author}{\bibinfo{person}{Nilesh Tripuraneni}, \bibinfo{person}{Lee Richardson}, \bibinfo{person}{Alexander D'Amour}, \bibinfo{person}{Jacopo Soriano}, {and} \bibinfo{person}{Steve Yadlowsky}.} \bibinfo{year}{2023}\natexlab{}.
\newblock \bibinfo{title}{Choosing a Proxy Metric from Past Experiments}.
\newblock
\newblock
\showeprint[arxiv]{2309.07893}~[stat.ME]


\bibitem[Urbano et~al\mbox{.}(2019)]%
        {Urbano2019}
\bibfield{author}{\bibinfo{person}{Juli\'{a}n Urbano}, \bibinfo{person}{Harlley Lima}, {and} \bibinfo{person}{Alan Hanjalic}.} \bibinfo{year}{2019}\natexlab{}.
\newblock \showarticletitle{Statistical Significance Testing in Information Retrieval: An Empirical Analysis of Type I, Type II and Type III Errors}. In \bibinfo{booktitle}{\emph{Proc. of the 42nd International ACM SIGIR Conference on Research and Development in Information Retrieval}} \emph{(\bibinfo{series}{SIGIR'19})}. \bibinfo{publisher}{ACM}, \bibinfo{pages}{505–514}.
\newblock
\showISBNx{9781450361729}
\urldef\tempurl%
\url{https://doi.org/10.1145/3331184.3331259}
\showDOI{\tempurl}


\bibitem[Ustimenko and Prokhorenkova(2020)]%
        {Ustimenko2020}
\bibfield{author}{\bibinfo{person}{Aleksei Ustimenko} {and} \bibinfo{person}{Liudmila Prokhorenkova}.} \bibinfo{year}{2020}\natexlab{}.
\newblock \showarticletitle{{S}tochastic{R}ank: Global Optimization of Scale-Free Discrete Functions}. In \bibinfo{booktitle}{\emph{Proc. of the 37th International Conference on Machine Learning}} \emph{(\bibinfo{series}{ICML '20'}, Vol.~\bibinfo{volume}{119})}. \bibinfo{publisher}{PMLR}, \bibinfo{pages}{9669--9679}.
\newblock
\urldef\tempurl%
\url{https://proceedings.mlr.press/v119/ustimenko20a.html}
\showURL{%
\tempurl}


\bibitem[Wald(1945)]%
        {Wald1945}
\bibfield{author}{\bibinfo{person}{Abraham Wald}.} \bibinfo{year}{1945}\natexlab{}.
\newblock \showarticletitle{{Sequential Tests of Statistical Hypotheses}}.
\newblock \bibinfo{journal}{\emph{The Annals of Mathematical Statistics}} \bibinfo{volume}{16}, \bibinfo{number}{2} (\bibinfo{year}{1945}), \bibinfo{pages}{117 -- 186}.
\newblock
\urldef\tempurl%
\url{https://doi.org/10.1214/aoms/1177731118}
\showDOI{\tempurl}


\bibitem[Wang et~al\mbox{.}(2022)]%
        {Wang2022}
\bibfield{author}{\bibinfo{person}{Yuyan Wang}, \bibinfo{person}{Mohit Sharma}, \bibinfo{person}{Can Xu}, \bibinfo{person}{Sriraj Badam}, \bibinfo{person}{Qian Sun}, \bibinfo{person}{Lee Richardson}, \bibinfo{person}{Lisa Chung}, \bibinfo{person}{Ed~H. Chi}, {and} \bibinfo{person}{Minmin Chen}.} \bibinfo{year}{2022}\natexlab{}.
\newblock \showarticletitle{Surrogate for Long-Term User Experience in Recommender Systems}. In \bibinfo{booktitle}{\emph{Proc. of the 28th ACM SIGKDD Conference on Knowledge Discovery and Data Mining}} (Washington DC, USA) \emph{(\bibinfo{series}{KDD '22})}. \bibinfo{publisher}{ACM}, \bibinfo{pages}{4100–4109}.
\newblock
\showISBNx{9781450393850}
\urldef\tempurl%
\url{https://doi.org/10.1145/3534678.3539073}
\showDOI{\tempurl}


\bibitem[WELCH(1947)]%
        {Welch1947}
\bibfield{author}{\bibinfo{person}{Bernard~Lewis WELCH}.} \bibinfo{year}{1947}\natexlab{}.
\newblock \showarticletitle{{The Generalization of ‘Student's’ Problem when Several Different Population Variances are Involved}}.
\newblock \bibinfo{journal}{\emph{Biometrika}} \bibinfo{volume}{34}, \bibinfo{number}{1-2} (\bibinfo{date}{01} \bibinfo{year}{1947}), \bibinfo{pages}{28--35}.
\newblock
\showISSN{0006-3444}
\urldef\tempurl%
\url{https://doi.org/10.1093/biomet/34.1-2.28}
\showDOI{\tempurl}


\bibitem[Xie and Aurisset(2016)]%
        {Xie2016}
\bibfield{author}{\bibinfo{person}{Huizhi Xie} {and} \bibinfo{person}{Juliette Aurisset}.} \bibinfo{year}{2016}\natexlab{}.
\newblock \showarticletitle{Improving the Sensitivity of Online Controlled Experiments: Case Studies at Netflix}. In \bibinfo{booktitle}{\emph{Proc. of the 22nd ACM SIGKDD International Conference on Knowledge Discovery and Data Mining}} \emph{(\bibinfo{series}{KDD '16})}. \bibinfo{publisher}{ACM}, \bibinfo{pages}{645–654}.
\newblock
\showISBNx{9781450342322}
\urldef\tempurl%
\url{https://doi.org/10.1145/2939672.2939733}
\showDOI{\tempurl}


\bibitem[Yue et~al\mbox{.}(2010)]%
        {Yue2010}
\bibfield{author}{\bibinfo{person}{Yisong Yue}, \bibinfo{person}{Yue Gao}, \bibinfo{person}{Oliver Chapelle}, \bibinfo{person}{Ya Zhang}, {and} \bibinfo{person}{Thorsten Joachims}.} \bibinfo{year}{2010}\natexlab{}.
\newblock \showarticletitle{Learning More Powerful Test Statistics for Click-Based Retrieval Evaluation}. In \bibinfo{booktitle}{\emph{Proc. of the 33rd International ACM SIGIR Conference on Research and Development in Information Retrieval}} \emph{(\bibinfo{series}{SIGIR '10})}. \bibinfo{publisher}{ACM}, \bibinfo{pages}{507–514}.
\newblock
\showISBNx{9781450301534}
\urldef\tempurl%
\url{https://doi.org/10.1145/1835449.1835534}
\showDOI{\tempurl}


\end{thebibliography}
